\ifdefined\pdfobjcompresslevel\pdfobjcompresslevel=0\fi
\documentclass{article}
\def\opdiadincludeappendix{}
\usepackage{iclr2027_conference,times}

\usepackage{amsmath,amsfonts,bm}

\def\eqref#1{equation~\ref{#1}}
\def\1{\bm{1}}

\DeclareMathAlphabet{\mathsfit}{\encodingdefault}{\sfdefault}{m}{sl}
\SetMathAlphabet{\mathsfit}{bold}{\encodingdefault}{\sfdefault}{bx}{n}

\usepackage{hyperref}
\usepackage{url}
\usepackage{graphicx}
\usepackage{float}
\usepackage{wrapfig}
\usepackage{algorithm}
\usepackage{algpseudocode}
\usepackage{booktabs}
\usepackage{multirow}
\usepackage{amsmath}
\usepackage{amssymb}
\usepackage{amsfonts}
\usepackage{amsthm}
\usepackage{microtype}
\usepackage{nicefrac}
\usepackage{xspace}

\usepackage{xcolor}
\usepackage{colortbl}

\newcommand{\method}{MOPDA\xspace}
\newcommand{\mopdasuppref}[1]{\hyperref[#1]{Appendix~\ref*{#1}}}

\title{MOPDA: Mixed-Trajectory On-Policy Distillation for Language-Guided Industrial Anomaly Detection}

\newif\ificlrsubmission
\iclrsubmissionfalse
\iclrfinalcopy

\author{
\textbf{Shuimu Chen}$^{*,1}$,
\textbf{Jing Jin}$^{*,1}$\\
Nan Su$^{1}$,
Hongbo Xu$^{2}$,
Zebang Cheng$^{2,3}$,
Wenming Yang$^{1}$,
Fei Ma$^{\dagger,2}$,
Guijin Wang$^{\dagger,1}$ \\[1mm]
$^{1}$ Tsinghua University \\
$^{2}$ Guangdong Laboratory of Artificial Intelligence and Digital Economy (SZ) \\
$^{3}$ Shenzhen University
}

\begin{document}

\maketitle
\lhead{}% No conference running header.
\begingroup
\renewcommand{\thefootnote}{\fnsymbol{footnote}}
\footnotetext[1]{Equal contribution.}
\footnotetext[2]{Corresponding authors.}
\endgroup

% Content first. Pagination is deferred until the text settles, so there are
% no forced page breaks between blocks for now.
\begin{abstract}
Large vision-language models (LVLMs) have shown strong potential for industrial
anomaly detection (IAD) by providing image-level anomaly judgments and
interpretable reasoning. However, reliably translating generated judgments into
precise pixel-level localization remains challenging. We propose
\textbf{M}ixed-Trajectory \textbf{O}n-\textbf{P}olicy \textbf{D}istillation for
Language-Guided Industrial \textbf{A}nomaly Detection (\textbf{\method}), the
first framework to introduce on-policy self-distillation into LVLM-based IAD.
For judgment learning, \method introduces \textbf{Mixed-Trajectory Supervision},
combining student-generated on-policy trajectories with evidence-conditioned
teacher trajectories under a shared token-level distillation objective. Student
trajectories preserve supervision on deployment-relevant response paths, while
teacher trajectories provide complementary evidence-conditioned supervision.
For dense localization, \textbf{Language-guided Visual Anchoring} uses the final
judgment as a compact semantic condition to construct image-specific normal and
abnormal anchors, which are contrasted with dense visual features to produce
anomaly maps. This keeps language as semantic guidance while grounding
pixel-level responses in visual evidence. Under a strict cross-dataset zero-shot
protocol on five IAD benchmarks, \method outperforms the evaluated LVLM-based
baselines on most detection, localization, and judgment metrics while remaining
competitive with CLIP-based methods. Ablations further validate both
mixed-trajectory supervision and final-judgment conditioning.
\end{abstract}

\section{Introduction}

Industrial anomaly detection (IAD) aims to identify defective products and
highlight anomalous regions, which is essential for visual quality inspection.
Existing IAD methods have established strong dense detection baselines through
CLIP-style representations, prompt designs, and visual foundation features
\citep{winclip,anomalyclip,adaclip,musc,visualad}. Though these methods have
made substantial progress, they do not generate natural-language explanations
of what defects are present, where they occur, or what visual evidence supports
their predictions.

Recent LVLM-based IAD methods introduce generative vision-language models into
IAD, enabling models to
understand industrial images, answer defect-related questions, and generate
image-specific anomaly judgments and explanations
\citep{anomalygpt,myriad,anomalyov,anomalyr1,iadr1}. However, reliably
converting these generated language judgments into precise pixel-level anomaly
maps remains underexplored in LVLM-based IAD.

\begin{figure*}[t]
\centering
\includegraphics[width=\textwidth]{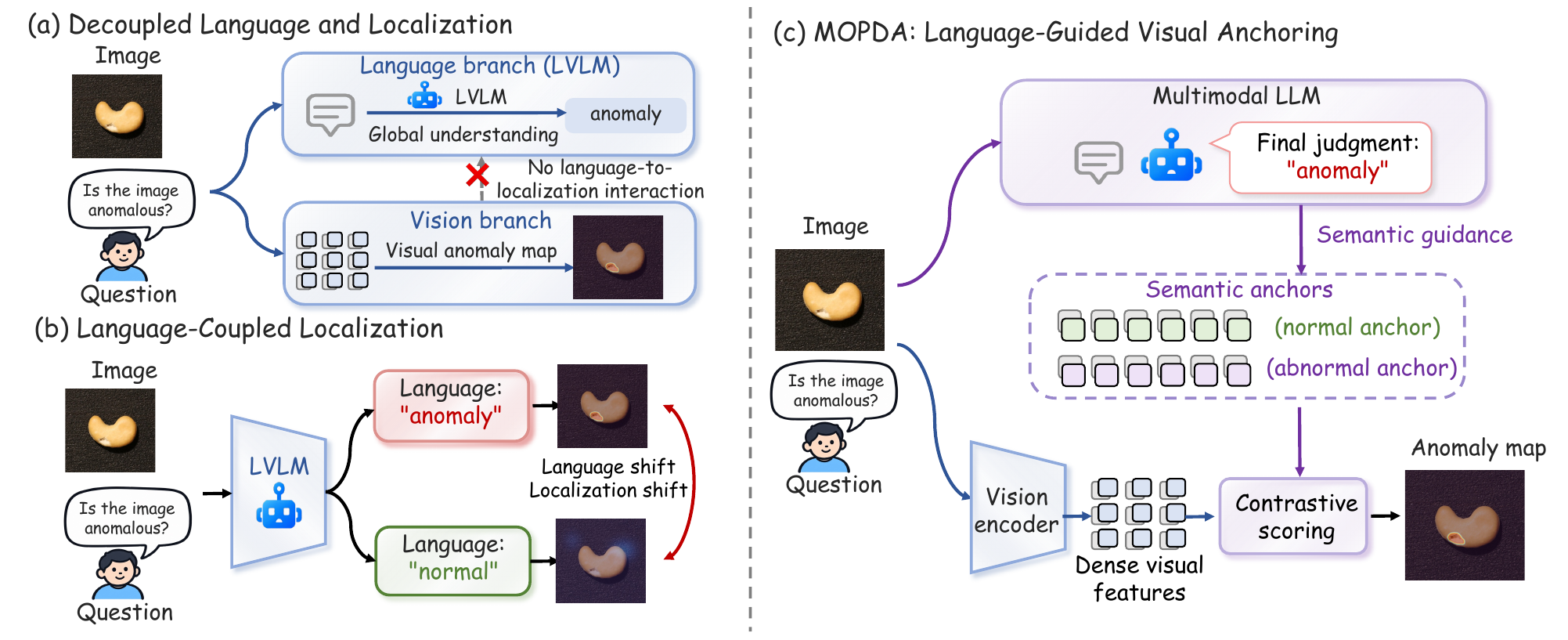}
\caption{
\textbf{Localization pathways in LVLM-based IAD.}
(a) Decoupled methods separate language generation from anomaly localization.
(b) Language-coupled methods derive localization from language-side signals,
making localization sensitive to shifts in these signals.
(c) MOPDA uses the final judgment as semantic guidance for normal/abnormal
anchors, while dense visual features ground contrastive pixel-level anomaly scoring.
}
\label{fig:localization-pathways}
\end{figure*}

This limitation appears in two forms in current localization pathways, as
illustrated in Figure~\ref{fig:localization-pathways}. In decoupled designs
(Figure~\ref{fig:localization-pathways}a), language generation and anomaly
localization follow separate pathways, so the generated language does not
directly inform heatmap prediction \citep{anomalygpt,myriad}. In
language-coupled designs (Figure~\ref{fig:localization-pathways}b), localization
is derived more directly from language-side grounding signals, such as
segmentation or semantic-anchor tokens, mask text, box-like outputs, and
reasoning-token attention \citep{agvas,vmad,real,omniad}. Consequently, changes
or errors in these language-side signals can propagate to the localization
output, making the predicted heatmap sensitive to language-side variations.
This motivates the design in Figure~\ref{fig:localization-pathways}c, where the
final judgment provides compact semantic guidance for localization, while
pixel-level anomaly scores remain grounded primarily in dense visual features.

Furthermore, using generated judgments to guide localization introduces a
distinct training challenge. Since the judgment is produced by the model itself
and then used as a semantic condition for localization, its training should
satisfy two requirements. First, supervision should target the model's own
generation trajectory, which is the trajectory used to produce the judgment at
deployment. Second, supervision should be fine-grained so that the judgment
captures defect evidence rather than merely optimizing a final-answer reward.

Existing LVLM training paradigms each satisfy only one side of these requirements.
Standard SFT provides dense token-level supervision, but learns from fixed
offline responses and therefore does not supervise the states visited by the
model at deployment. Reinforcement learning with verifiable rewards (RLVR),
in contrast, samples on-policy rollouts but typically relies on sparse
outcome-level signals, such as answer correctness, format, or coarse localization
\citep{deepseekmath,anomalyr1,iadr1}. Thus, SFT is fine-grained but off-policy,
whereas RLVR is on-policy but sparse. On-policy distillation bridges this gap by
applying token-level teacher supervision to student-generated on-policy rollouts,
combining on-policy training with fine-grained feedback
\citep{gkd,minillm,opsd}. However, in IAD, subtle defect evidence can lead the
current student to incomplete or incorrect judgment trajectories, limiting
student-only on-policy distillation to response paths sampled from the current
policy.

Motivated by this limitation, we propose
\textbf{M}ixed-Trajectory \textbf{O}n-\textbf{P}olicy
\textbf{D}istillation for Language-Guided Industrial
\textbf{A}nomaly Detection (\textbf{\method}),
which is the first framework to introduce on-policy self-distillation into
LVLM-based IAD.
For judgment learning, \method introduces
\textbf{Mixed-Trajectory Supervision}, which combines student on-policy
rollouts with evidence-conditioned teacher rollouts.
The student trajectories preserve supervision on deployment-relevant states,
while the teacher trajectories complement them with evidence-conditioned
judgment paths that the current student may not reliably generate.

\begin{figure*}[t]
\centering
\includegraphics[width=\textwidth]{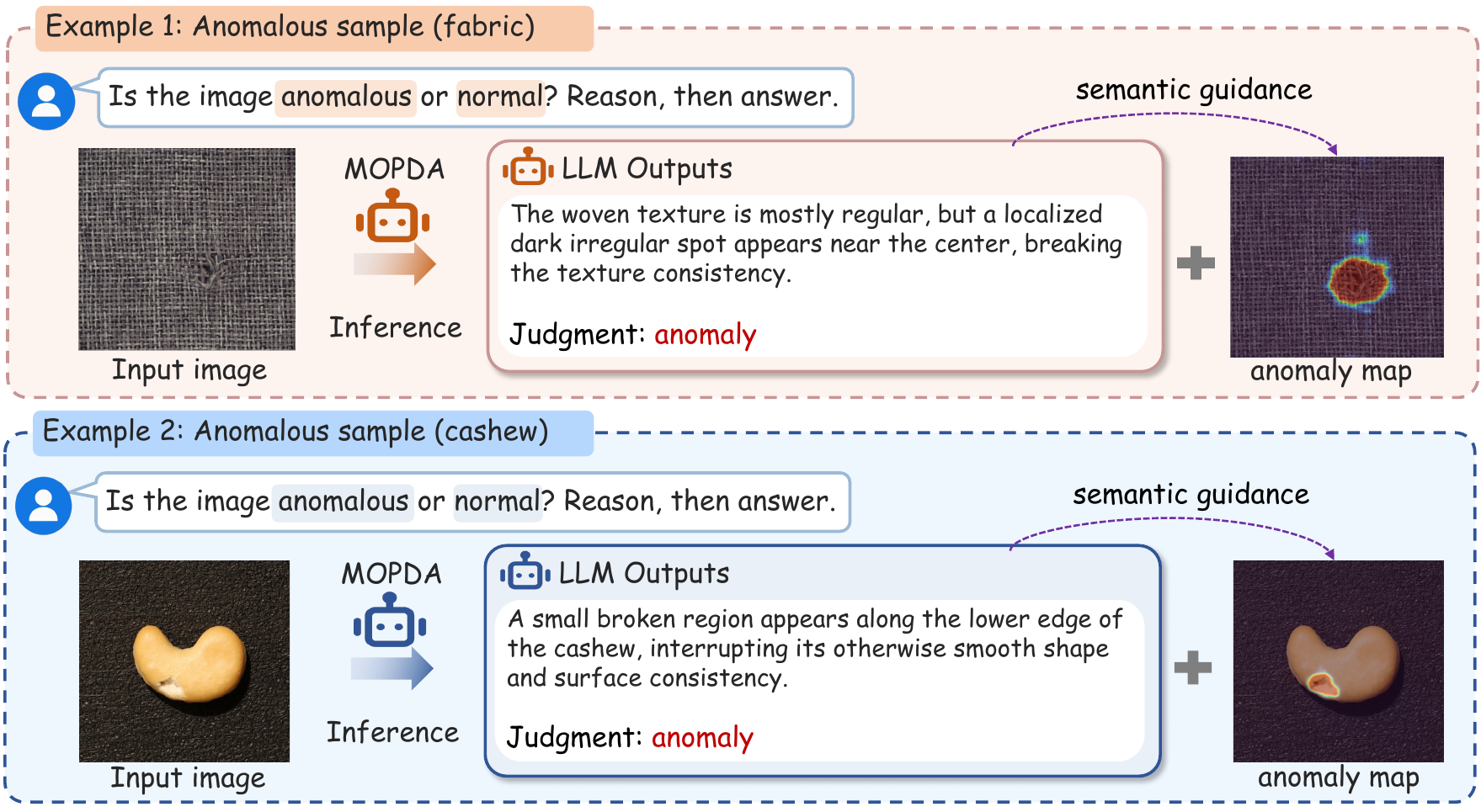}
\caption{\textbf{Language-guided anomaly localization.}
MOPDA correctly judges both the fabric (top) and cashew (bottom) as anomalous.
Their anomaly maps highlight the visible defects, with the final judgment
providing semantic guidance.}
\label{fig:qualitative}
\end{figure*}

Furthermore, we design \textbf{Language-guided Visual Anchoring} as a concrete
mechanism for this guidance. Instead of using the full chain-of-thought as the
localization context, \method keeps only the final anomaly judgment and
uses a judgment reforward to re-encode the image, question, and judgment into
normal and abnormal anchor representations. A contrastive heatmap head then
scores dense visual patch features against these anchors to produce the anomaly map.
This design keeps pixel-level anomaly responses grounded in visual features
rather than directly determined by the generated judgment.
Figure~\ref{fig:qualitative} provides an output-level illustration of this behavior:
in both examples, the generated reasoning identifies the visible defect, while
the corresponding anomaly map highlights the same spatial region, showing
qualitative consistency between language judgment and dense localization.
Experiments on five IAD benchmarks validate the effectiveness of our framework.

Our contributions are summarized as follows.
\begin{itemize}
\item[1.] We present \textbf{\method}, the first framework to introduce
on-policy self-distillation into LVLM-based IAD, and develop \textbf{Mixed-Trajectory
Supervision} that combines student-generated on-policy trajectories with
evidence-conditioned teacher-generated trajectories.

\item[2.] We design \textbf{Language-guided Visual Anchoring}, which
conditions normal and abnormal anchors on the final judgment while grounding
pixel-level anomaly scores in visual features.

\item[3.] Experiments on five IAD benchmarks demonstrate competitive
visual anomaly detection against CLIP-based methods and improved
detection, localization, and judgment performance over the evaluated
LVLM baselines.
\end{itemize}

\section{Related Work}
\label{sec:related}

\subsection{Industrial Anomaly Detection}

Zero-shot industrial anomaly detection commonly builds on pretrained
vision-language or visual representations. WinCLIP~\citep{winclip} derives
image- and pixel-level anomaly scores from local CLIP features and semantic
prompts, AnomalyCLIP~\citep{anomalyclip} learns object-agnostic prompts for
cross-category generalization, and VisualAD~\citep{visualad} emphasizes dense
visual matching with reduced dependence on language. These methods provide
strong anomaly detection and localization, but do not generate image-specific
language judgments.

Recent LVLM-based methods extend IAD toward language-based anomaly understanding.
AnomalyGPT~\citep{anomalygpt} uses a separate localization pathway, Myriad~\citep{myriad}
incorporates visual experts, ReAL~\citep{real} derives dense responses from
reasoning attention, and AG-VAS~\citep{agvas} introduces semantic anchor tokens
with anchor-guided mask decoding. In contrast, \method uses the generated language judgment as semantic guidance while preserving a visual-feature-grounded localization pathway.

In a broader multimodal setting, HUG-VIS~\citep{ma2026hug} benchmarks
human-centered understanding and generation across video, audio, and language.

\subsection{On-Policy Distillation}

On-policy distillation provides token-level teacher supervision on
student-generated trajectories, reducing training--inference mismatch
\citep{gkd,minillm}. Its self-distillation form obtains supervision from the
same model under privileged contexts~\citep{opsd}, and recent work extends this
paradigm to multimodal reasoning. In industrial anomaly detection,
ADOPD~\citep{adopd} uses privileged references to improve query-only industrial
reasoning.

\method differs in two respects. First, it introduces
\textbf{Mixed-Trajectory Supervision}, augmenting student-generated on-policy
trajectories with evidence-conditioned teacher-generated trajectories under a
shared token-level distillation objective. Second, the learned judgment is not
only a reasoning output, but is further used to guide dense anomaly localization
through Language-guided Visual Anchoring.
\section{Method}

\subsection{Setting and Framework Overview}

We study IAD under a cross-dataset setting, training on labeled source data
and evaluating on unseen target domains. For each sample, the observable
input is $x_i=(x_i^{\mathrm{img}},x_i^{\mathrm{text}})$, where
$x_i^{\mathrm{img}}$ is an industrial image and $x_i^{\mathrm{text}}$ is a
defect-judgment question. The source training set is
\begin{equation}
\mathcal{D}_{\mathrm{src}}
=
\{(x_i^{\mathrm{img}},x_i^{\mathrm{text}},a_i,M_i,r_i)\}_{i=1}^{N},
\end{equation}
where $a_i$ is the image-level anomaly label, $M_i$ is the pixel-level anomaly
mask, and $r_i$ is a defect-analysis reference containing a verified judgment
and supporting evidence. These annotations and references are used only
during source-domain training. At target-domain inference, the model receives
only the image and question and produces a language response and a pixel-level
anomaly map.

\begin{figure}[t]
\centering
\includegraphics[width=\linewidth]{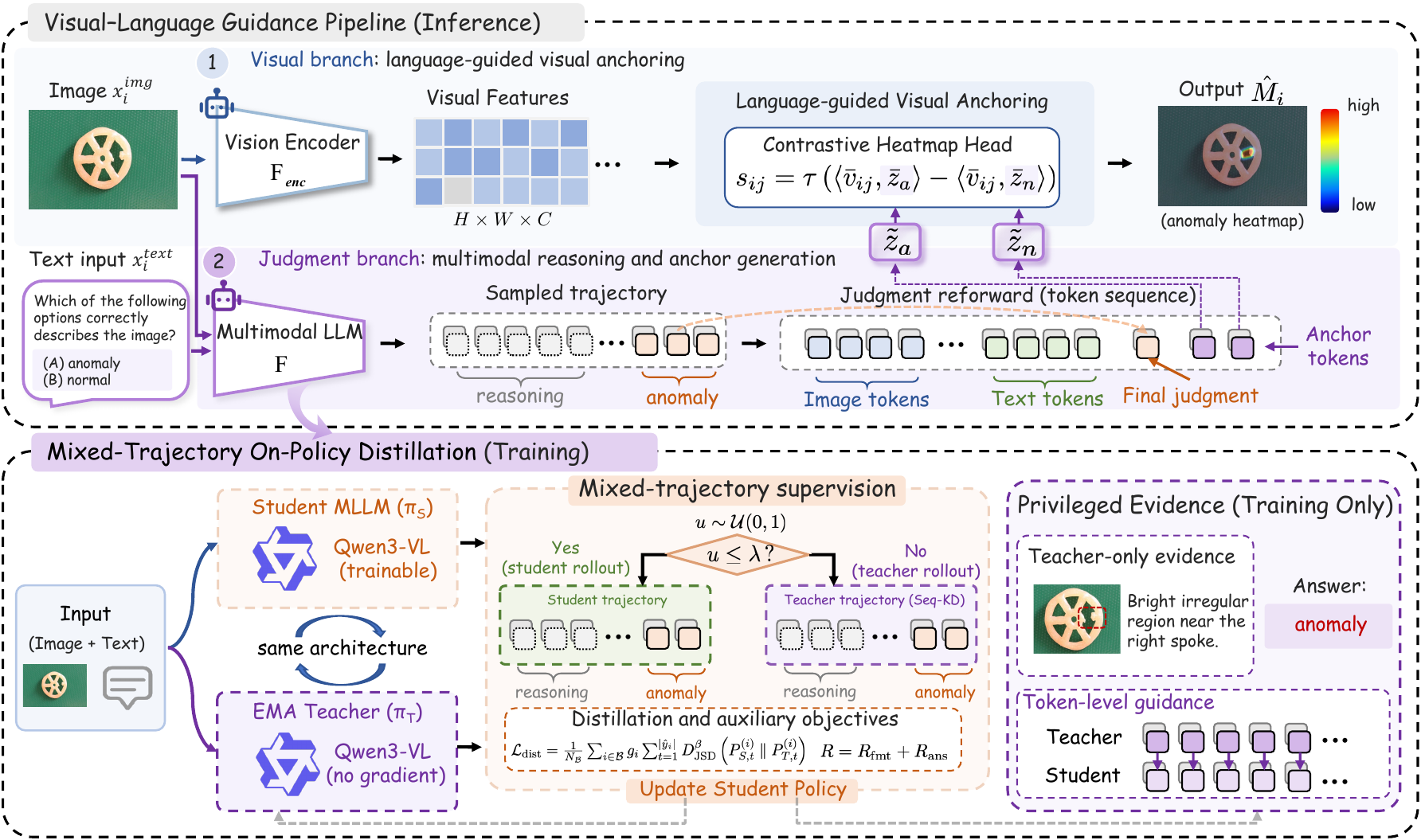}
\caption{\textbf{Overview of \method.} Mixed-Trajectory Supervision uses
student-generated on-policy trajectories and evidence-conditioned
teacher-generated trajectories (Seq-KD). The student learns from both through
a shared token-level distillation objective, while only the teacher observes
privileged evidence during training. Language-guided Visual Anchoring uses a
judgment reforward to contextualize the base abnormal and normal anchor tokens
$z_a,z_n$ into image-specific anchor representations for dense anomaly
localization.}
\label{fig:pipeline}
\end{figure}

As shown in Figure~\ref{fig:pipeline}, \method uses the final language
judgment to contextualize abnormal and normal anchor tokens through a judgment
reforward. A contrastive heatmap head scores dense visual features against
the resulting anchor representations to produce the anomaly map, whose spatial
aggregation gives the image-level anomaly score. During training,
Mixed-Trajectory Supervision combines student- and teacher-generated
trajectories. Privileged defect evidence is available only to the teacher,
while source labels and masks supervise anomaly prediction. Only the student
is retained at inference.

\subsection{Mixed-Trajectory On-Policy Distillation}

The student receives only the image and question, while the teacher additionally
observes the training-only defect-analysis reference $r_i$. Following
privileged-context self-distillation~\citep{opsd}, their contexts are
\begin{equation}
c_i^S
=
\mathrm{Prompt}_S(x_i^{\mathrm{img}},x_i^{\mathrm{text}}),
\qquad
c_i^T
=
\mathrm{Prompt}_T(x_i^{\mathrm{img}},x_i^{\mathrm{text}},r_i).
\label{eq:distillation-contexts}
\end{equation}
The teacher shares the student architecture and is maintained by
EMA~\citep{tarvainen2017mean}, as detailed in
\mopdasuppref{app:distillation-details}.

To construct Mixed-Trajectory Supervision, we sample the trajectory source
for each minibatch with $b\sim\mathrm{Bernoulli}(\lambda)$:
\begin{equation}
\hat y_i\sim
\begin{cases}
\pi_\theta(\cdot\mid c_i^S), & b=1,\\
\pi_{\bar\theta}(\cdot\mid c_i^T), & b=0.
\end{cases}
\label{eq:mixed-trajectories}
\end{equation}
Student-generated trajectories retain supervision on response paths explored
by the current policy and used at deployment. Teacher-generated trajectories,
which form the Seq-KD branch, provide complementary evidence-conditioned
judgment paths beyond those sampled from the student policy.

Both trajectory sources are optimized with the same token-level distillation
objective. For a sampled response $\hat y_i$, the student and teacher evaluate
the same response prefix under their respective contexts. Let
$P_{S,t}^{(i)}$ and $P_{T,t}^{(i)}$ denote their corresponding next-token
distributions at step $t$. We minimize
\begin{equation}
\mathcal L_{\mathrm{dist}}
=
\frac{1}{N_{\mathcal B}}
\sum_{i\in\mathcal B}
g_i
\sum_{t=1}^{|\hat y_i|}
D_{\mathrm{JSD}}^\beta
\left(
P_{S,t}^{(i)}
\|
P_{T,t}^{(i)}
\right),
\label{eq:mixed-distillation-loss}
\end{equation}
where $g_i=1$ for student-generated trajectories, while for teacher-generated
trajectories $g_i$ acts as a validity gate that retains only responses with the
required output format and a final judgment matching the source label.
$N_{\mathcal B}$ denotes the total number of sampled response tokens in the
minibatch. The exact next-token distributions, generalized Jensen--Shannon
divergence, validity criterion, and normalization are provided in
\mopdasuppref{app:distillation-details}.

\subsection{Language-guided Visual Anchoring}

Language-guided Visual Anchoring uses the final judgment $\hat a_i$, extracted
from the \texttt{<answer></answer>} span of response $\hat y_i$, as a compact
semantic condition for dense localization. The same judgment is returned as
the language output and is also used to construct image-specific abnormal and
normal anchor representations. Figure~\ref{fig:qualitative} illustrates this
inference-time behavior: the model outputs the final judgment while the same
judgment conditions the anchors used for dense anomaly localization.
Table~\ref{tab:anchor} evaluates this design against alternative localization
interfaces.

Specifically, we append two base anchor tokens, denoted by $z_a$ and $z_n$,
representing abnormal and normal semantics, respectively. A judgment reforward
with the same LVLM processes the image, question, final judgment $\hat a_i$,
and the two anchor tokens:
\begin{equation}
(\tilde{z}_{i,a},\tilde{z}_{i,n})
=
f^{\mathrm{ref}}_{\theta}
\left(
x_i^{\mathrm{img}},
x_i^{\mathrm{text}},
\hat{a}_i,
z_a,z_n
\right).
\label{eq:judgment_reforward}
\end{equation}
Here, $\tilde{z}_{i,a}$ and $\tilde{z}_{i,n}$ are the contextualized hidden
states of the abnormal and normal anchor tokens after reforwarding. They
therefore encode the image content, question context, and final-judgment
semantics for sample $i$.

The contextualized anchor states are mapped into a shared localization space
by two learnable projection heads $g_a$ and $g_n$ for the abnormal and normal
anchors, respectively, and are then L2-normalized:
\begin{equation}
\bar z_{i,a}
=
\frac{g_a(\tilde z_{i,a})}
{\|g_a(\tilde z_{i,a})\|_2},
\qquad
\bar z_{i,n}
=
\frac{g_n(\tilde z_{i,n})}
{\|g_n(\tilde z_{i,n})\|_2}.
\label{eq:anchor_projection}
\end{equation}

Let $v_{i,p}$ denote the dense visual feature of patch $p$ in sample $i$.
A lightweight visual refinement module maps these features into the same
localization space as the anchors, followed by L2 normalization to obtain
$\bar v_{i,p}$. The anomaly logit for each patch is then computed by
contrasting its similarity to the abnormal and normal anchors:
\begin{equation}
s_{i,p}
=
\tau
\left(
\langle \bar v_{i,p}, \bar z_{i,a} \rangle
-
\langle \bar v_{i,p}, \bar z_{i,n} \rangle
\right),
\label{eq:contrastive_heatmap}
\end{equation}
where the learnable temperature $\tau$ multiplies the similarity difference.
Upsampling the spatial logits
and applying a sigmoid yields the pixel-level anomaly map $\hat M_i$.
The detailed head architecture, projection dimensions, normalization, and
image-score aggregation are provided in
\mopdasuppref{app:anchoring-details}.

\subsection{Training Objective}

Training jointly optimizes judgment learning and anomaly localization. We
combine mixed-trajectory distillation with a lightweight outcome objective on
student-generated trajectories and supervised anomaly localization:
\begin{equation}
\mathcal{L}
=
\mathcal{L}_{\mathrm{dist}}
+
\lambda_R\mathcal{L}_{\mathrm{reward}}
+
\lambda_H\mathcal{L}_{\mathrm{map}}.
\label{eq:joint-training-loss}
\end{equation}
Here, $\mathcal{L}_{\mathrm{dist}}$ provides token-level supervision for both
student- and teacher-generated trajectories. On student rollouts,
$\mathcal{L}_{\mathrm{reward}}$ provides lightweight outcome supervision for
response-format validity and final-judgment correctness, while it is disabled
for teacher-generated Seq-KD trajectories. $\mathcal{L}_{\mathrm{map}}$
combines image-level anomaly supervision, pixel-level localization supervision,
and normal--abnormal anchor separation. Together, these objectives couple
fine-grained judgment learning with dense anomaly localization. Exact loss
definitions, component weights, and optimization details are provided in
\mopdasuppref{app:training-objective}.

\ifdefined\opdiadincludeappendix
The complete training procedure is summarized in
Algorithm~\ref{alg:mopda-training}.
\else
The complete training procedure is summarized in the supplementary material.
\fi

\definecolor{tablegroup}{HTML}{E9EDF2}
\definecolor{tablefamily}{HTML}{F5F5F5}
\definecolor{tableoracle}{HTML}{E8E8E8}
\definecolor{tablebest}{HTML}{F5B7BF}
\definecolor{tablesecond}{HTML}{D5DEFF}
\definecolor{posfill}{RGB}{221,233,245} % very soft blue
\definecolor{negfill}{RGB}{245,226,204} % very soft peach
\newcommand{\bestfill}[1]{\begingroup\setlength{\fboxsep}{0.45pt}\colorbox{tablebest}{\strut\textbf{#1}}\endgroup}
\newcommand{\secondfill}[1]{\begingroup\setlength{\fboxsep}{0.45pt}\colorbox{tablesecond}{\strut #1}\endgroup}
\newcommand{\changefill}[2]{\begingroup\setlength{\fboxsep}{0.45pt}\colorbox{#1}{\makebox[4.5em][c]{\strut #2}}\endgroup}
\newcommand{\positivefill}[1]{\changefill{posfill}{#1}}
\newcommand{\negativefill}[1]{\changefill{negfill}{#1}}
\newcommand{\metriccell}[3]{#1\,/\,#2\,/\,#3}
\newcommand{\nodata}{--}

\section{Experiments}

\subsection{Experimental setup}
\paragraph{Datasets and protocol.}
We evaluate \method under a strict cross-dataset zero-shot protocol
on five benchmarks: MVTec-AD~\citep{mvtecad}, VisA~\citep{visa},
SDD/KolektorSDD~\citep{kolektorsdd}, WFDD~\citep{wfdd}, and
KSDD2~\citep{ksdd2}. For MVTec-AD, we train on VisA and
GoodsAD~\citep{goodsad}. For VisA, we train on MVTec-AD and
GoodsAD.
For SDD, WFDD, and KSDD2, we train on MVTec-AD and GoodsAD.
The complete assignments are tabulated in
\mopdasuppref{app:data-construction}.
\paragraph{Model and training.}
We use Qwen3-VL-8B-Instruct~\citep{qwen3vl} as the backbone and fine-tune
all model parameters, including the vision encoder. Models are trained for
30 epochs on 8 A100 GPUs using DeepSpeed ZeRO-2, with an effective batch
size of 32, a learning rate of $1\times10^{-6}$, and a warmup ratio of 0.03.
For \method, we use an EMA decay of 0.999, set the generalized JSD parameter
to $\beta=0.5$, and sample student-generated trajectories with probability
$\lambda=0.8$. The auxiliary reward loss is weighted by 0.1.
Additional optimization details are provided in
\mopdasuppref{app:training-objective}.

\paragraph{Metrics.}
We report image-level and pixel-level AUROC, AP, and F1,
together with binary judgment accuracy (QA/Acc.), in percent.
All \method anomaly metrics are evaluated on the complete
test set with the full-resolution protocol, while QA/Acc.\
is computed from the final binary language judgment.

\subsection{Main results}
\begin{wrapfigure}{r}{0.58\linewidth}
  \centering
  \includegraphics[width=\linewidth]{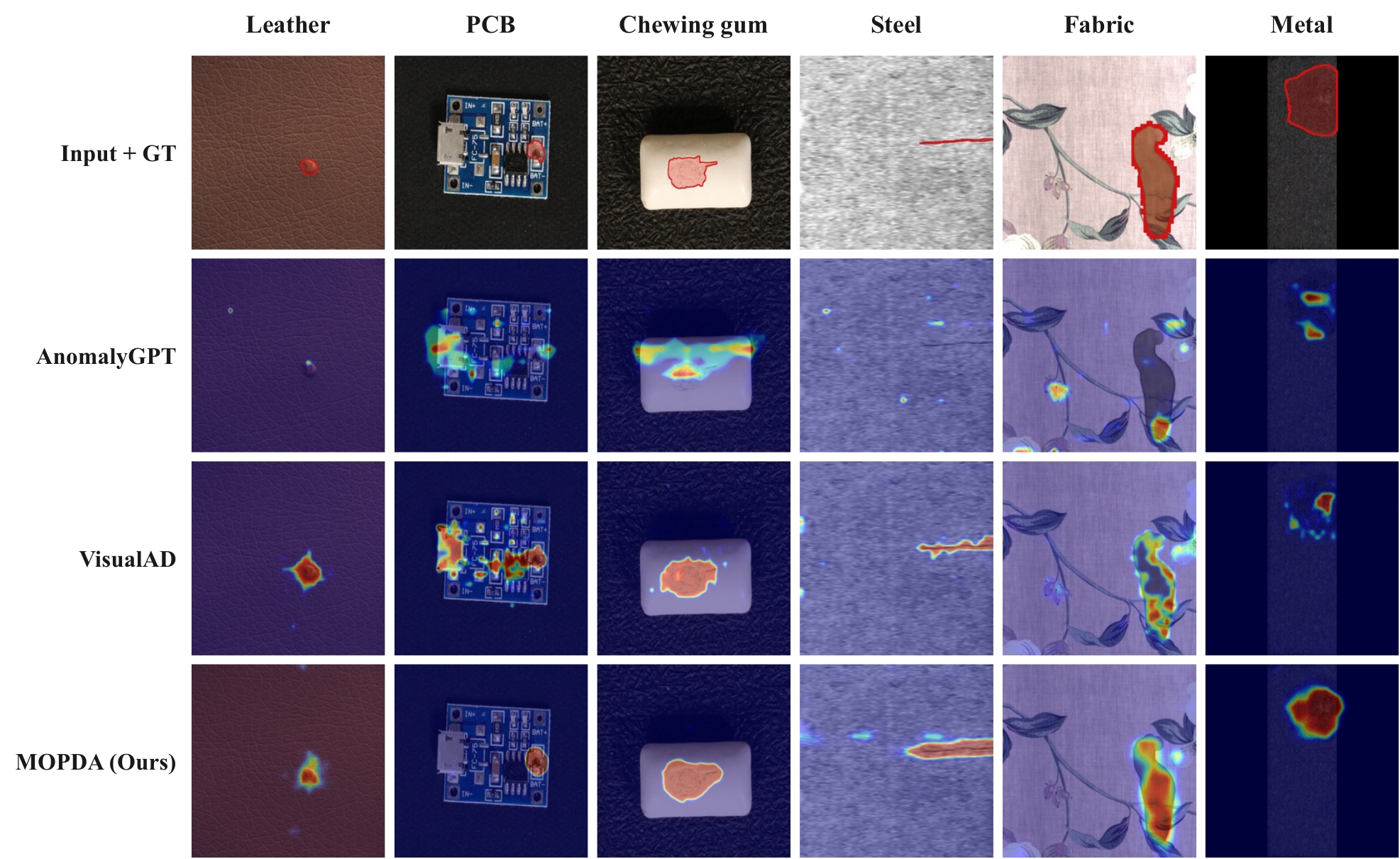}
  \caption{\textbf{Cross-benchmark qualitative comparison.}
  Anomaly maps for six categories from five benchmarks. Ground-truth masks
  are overlaid in red.}
  \label{fig:cross-benchmark-qualitative}
\end{wrapfigure}
Table~\ref{tab:main} compares CLIP-based and LVLM-based methods
on five industrial anomaly detection benchmarks.
MVTec-AD and VisA cover diverse industrial objects and textures,
while SDD, WFDD, and KSDD2 extend the evaluation to surface
and fabric defects.
\method achieves competitive image-level discrimination and
pixel-level localization against CLIP-based methods, while
also producing explicit binary language judgments.
Compared with the evaluated LVLM baselines, it improves most
reported detection, localization, and judgment metrics.
Figure~\ref{fig:cross-benchmark-qualitative} further compares representative
localization results on six categories from the five benchmarks. AnomalyGPT
often produces sparse or off-target responses, while VisualAD sometimes
activates over broader or multiple regions. \method generally
concentrates its response on the annotated defects across object, texture, and
surface categories. Together with Table~\ref{tab:main}, these examples
illustrate how its competitive quantitative performance translates into
spatially coherent localization.

\begin{table}[t]
\centering
\caption{\textbf{Main comparison on five IAD benchmarks (\%).}
Image- and pixel-level cells report AUROC/AP/F1, and language judgment reports
QA/Acc. Best results are shaded pink and second-best results blue.
$^\dagger$ marks our reproduction.
``--'' denotes an unavailable result.}
\label{tab:main}
\scriptsize
\renewcommand{\arraystretch}{1.14}
\setlength{\tabcolsep}{3.2pt}
\resizebox{\linewidth}{!}{%
\begin{tabular}{lccccc}
\toprule
\textbf{Method} & \textbf{MVTec-AD} & \textbf{VisA} & \textbf{SDD} & \textbf{WFDD} & \textbf{KSDD2} \\
\midrule
\rowcolor{tablegroup}\multicolumn{6}{l}{\textbf{Image-level} \textit{(AUROC / AP / F1)}} \\
\midrule
\rowcolor{tablefamily}\multicolumn{6}{l}{\textit{CLIP-based}} \\
WinCLIP$^\dagger$ & \metriccell{70.70}{84.91}{86.46} & \metriccell{65.22}{70.94}{74.62} & \metriccell{69.72}{29.04}{36.67} & \metriccell{72.52}{74.42}{75.01} & \metriccell{83.28}{85.90}{76.40} \\
AnomalyCLIP$^\dagger$ & \metriccell{91.40}{96.13}{91.89} & \metriccell{79.25}{83.54}{79.56} & \metriccell{\secondfill{91.08}}{\secondfill{53.97}}{\secondfill{59.26}} & \metriccell{84.00}{\secondfill{86.42}}{\secondfill{83.21}} & \metriccell{91.52}{94.60}{\secondfill{90.06}} \\
VisualAD$^\dagger$ & \metriccell{\secondfill{92.22}}{\secondfill{96.69}}{\secondfill{92.34}} & \metriccell{\secondfill{84.58}}{\secondfill{87.33}}{\secondfill{82.45}} & \metriccell{83.10}{31.39}{47.25} & \metriccell{\secondfill{84.23}}{85.98}{\bestfill{83.82}} & \metriccell{\bestfill{98.42}}{\bestfill{98.65}}{\bestfill{94.57}} \\
\rowcolor{tablefamily}\multicolumn{6}{l}{\textit{LVLM-based}} \\
AnomalyGPT$^\dagger$ & \metriccell{71.43}{85.94}{85.95} & \metriccell{60.79}{68.20}{73.84} & \metriccell{64.72}{22.81}{31.47} & \metriccell{61.37}{69.10}{71.13} & \metriccell{87.30}{89.20}{80.11} \\
Myriad$^\dagger$ & \nodata & \metriccell{51.93}{57.88}{71.39} & \metriccell{64.68}{22.79}{31.63} & \metriccell{62.99}{68.95}{71.13} & \metriccell{87.31}{89.21}{80.10} \\
ReAL & \metriccell{79.80}{87.10}{--} & \nodata & \metriccell{81.40}{44.30}{--} & \metriccell{79.90}{78.80}{--} & \nodata \\
\textbf{MOPDA} & \metriccell{\bestfill{93.76}}{\bestfill{96.99}}{\bestfill{92.58}} & \metriccell{\bestfill{89.01}}{\bestfill{91.91}}{\bestfill{85.98}} & \metriccell{\bestfill{96.15}}{\bestfill{77.77}}{\bestfill{72.22}} & \metriccell{\bestfill{88.53}}{\bestfill{90.43}}{82.85} & \metriccell{\secondfill{95.48}}{\secondfill{96.30}}{88.70} \\
\midrule
\rowcolor{tablegroup}\multicolumn{6}{l}{\textbf{Pixel-level} \textit{(AUROC / AP / F1)}} \\
\midrule
\rowcolor{tablefamily}\multicolumn{6}{l}{\textit{CLIP-based}} \\
WinCLIP$^\dagger$ & \metriccell{80.93}{16.82}{23.20} & \metriccell{70.54}{6.32}{9.79} & \metriccell{72.18}{1.14}{6.19} & \metriccell{67.25}{9.05}{13.54} & \metriccell{96.79}{22.69}{31.05} \\
AnomalyCLIP$^\dagger$ & \metriccell{88.20}{33.56}{38.68} & \metriccell{93.94}{18.96}{27.32} & \metriccell{\secondfill{95.80}}{\secondfill{10.65}}{\secondfill{21.09}} & \metriccell{\secondfill{90.34}}{\bestfill{34.27}}{\bestfill{39.80}} & \metriccell{91.45}{45.33}{56.26} \\
VisualAD$^\dagger$ & \metriccell{\secondfill{90.78}}{\secondfill{41.30}}{\secondfill{43.71}} & \metriccell{\bestfill{95.74}}{\secondfill{28.47}}{\secondfill{35.09}} & \metriccell{94.55}{5.67}{12.51} & \metriccell{82.33}{18.77}{25.04} & \metriccell{\secondfill{98.21}}{\secondfill{60.42}}{\secondfill{59.69}} \\
\rowcolor{tablefamily}\multicolumn{6}{l}{\textit{LVLM-based}} \\
AnomalyGPT$^\dagger$ & \metriccell{77.61}{21.41}{28.76} & \metriccell{81.68}{6.64}{11.48} & \metriccell{62.33}{0.18}{1.86} & \metriccell{75.62}{7.25}{13.24} & \metriccell{90.70}{49.31}{51.44} \\
Myriad$^\dagger$ & \nodata & \metriccell{82.06}{4.12}{8.57} & \metriccell{62.34}{0.18}{1.87} & \metriccell{75.87}{8.28}{11.53} & \metriccell{90.70}{49.31}{51.45} \\
ReAL & \metriccell{75.60}{16.60}{--} & \nodata & \metriccell{82.30}{2.60}{--} & \metriccell{70.50}{7.40}{--} & \nodata \\
\textbf{MOPDA} & \metriccell{\bestfill{90.87}}{\bestfill{43.52}}{\bestfill{45.44}} & \metriccell{\secondfill{95.49}}{\bestfill{30.58}}{\bestfill{35.61}} & \metriccell{\bestfill{99.40}}{\bestfill{21.25}}{\bestfill{32.52}} & \metriccell{\bestfill{91.68}}{\secondfill{25.26}}{\secondfill{31.47}} & \metriccell{\bestfill{99.25}}{\bestfill{62.73}}{\bestfill{61.82}} \\
\midrule
\rowcolor{tablegroup}\multicolumn{6}{l}{\textbf{Language judgment} \textit{(Accuracy)}} \\
\midrule
AnomalyGPT$^\dagger$ & \secondfill{44.75} & 27.29 & 11.93 & 58.72 & 15.73 \\
Myriad$^\dagger$ & \nodata & \secondfill{55.50} & \secondfill{24.01} & \secondfill{60.00} & \secondfill{28.93} \\
\textbf{MOPDA} & \bestfill{71.42} & \bestfill{74.01} & \bestfill{83.71} & \bestfill{63.74} & \bestfill{87.50} \\
\bottomrule
\end{tabular}}
\end{table}

Figure~\ref{fig:qualitative} shows localized heatmaps for correctly judged
fabric and cashew anomalies. Table~\ref{tab:anchor} tests the role of the
judgment directly: ground-truth judgments improve all six anomaly metrics,
whereas incorrect judgments reduce them. Together, these results support
language as semantic guidance rather than a hard gate for dense anomaly
localization.

\subsection{Ablation studies}

\paragraph{Training paradigm.}
We compare four training paradigms under the same data, localization branch,
and trainable vision tower. SFT learns from fixed reference responses,
GRPO optimizes sparse outcome rewards on the model's own samples,
and OPSD applies privileged token-level distillation only to student-generated
on-policy trajectories. \method extends OPSD through \textbf{Mixed-Trajectory
Supervision} by additionally using evidence-conditioned teacher-generated
trajectories in the Seq-KD branch.

As shown in Table~\ref{tab:paradigm}, \method outperforms SFT on all six
anomaly-detection metrics, indicating the benefit of supervising
deployment-relevant trajectories rather than only imitating fixed references.
Compared with GRPO, \method sacrifices slightly on image-level AUROC/AP but
achieves substantially stronger pixel-level localization, showing the advantage
of dense token-level supervision over sparse outcome rewards for localization.

\paragraph{Effect of Mixed-Trajectory Supervision.}
Paired comparisons in Table~\ref{tab:paired-judgment} show that
Mixed-Trajectory Supervision improves judgment accuracy on three of the four
evaluated benchmarks, with substantially more errors corrected than correct
predictions regressed overall. The class-wise analysis further shows that the
correction patterns vary across target domains, with some datasets benefiting
more from improved anomalous-sample predictions and others from improved
normal-sample predictions. These results support the motivation for
Mixed-Trajectory Supervision. Student-only on-policy distillation is restricted
to response paths explored by the current policy, while evidence-conditioned
teacher trajectories provide complementary judgment paths and additional
supervision beyond student-generated trajectories.

\paragraph{Language-guided visual anchoring.}
Table~\ref{tab:anchor} analyzes \textbf{Language-guided Visual Anchoring} by
asking two questions: whether the localization branch should use the final
judgment as its compact semantic condition, and how strongly the heatmap depends
on the semantics of that judgment. The first block compares four MOPDA
variants: full-response conditioning, learnable anchors alone, no judgment
aggregation, and no language output. The no-aggregation variant still generates
the language response but prevents the final judgment from being aggregated
into the anchors. The no-language-output variant removes language output from
the localization path. The second block keeps the inspected image and sampled
response fixed, replacing only the judgment span fed to the judgment reforward
with an incorrect judgment or, as an analysis-only oracle, a GT judgment.

The interface controls identify the final judgment as the strongest deployable
condition: full responses introduce redundant noise, whereas learnable anchors
alone lack sample-specific semantics. The no-aggregation and no-language-output
controls further separate language generation from judgment aggregation:
keeping language output without
aggregating the final judgment remains close at the image level but drops on
dense localization metrics, and removing language output weakens localization
further. These results show that aggregating the final judgment into the visual
anchors improves localization beyond generating a language response alone.
An anchor-space visualization in
\mopdasuppref{app:anchor-space} illustrates the separation between normal and
abnormal anchors.
Judgment interventions support the same conclusion: GT judgments
improve all six reported anomaly metrics, whereas incorrect
judgments reduce them, confirming judgment semantics as an
effective localization signal.

\begin{table}[t]
\centering
\setlength{\belowcaptionskip}{6pt}
\caption{\textbf{Training-paradigm comparison on VisA.}
SFT, GRPO, OPSD, and \method use the same data, localization
branch, and trainable vision tower. Best results are shaded pink and second-best
results blue.}
\label{tab:paradigm}
\fontsize{8.2pt}{10pt}\selectfont
\renewcommand{\arraystretch}{1.12}
\begin{tabular*}{\linewidth}{@{\extracolsep{\fill}}lccccccc@{}}
\toprule
\multirow{2}{*}{Setting}
& \multicolumn{3}{c}{Image-level}
& \multicolumn{3}{c}{Pixel-level}
& \multicolumn{1}{c}{Language} \\
\cmidrule(lr){2-4}\cmidrule(lr){5-7}\cmidrule(lr){8-8}
& AUROC & AP & F1 & AUROC & AP & F1 & QA/Acc. \\
\midrule
SFT & 87.53 & 91.19 & 85.46 & 84.76 & 21.66 & 29.66 & \secondfill{71.88} \\
GRPO & \bestfill{89.68} & \bestfill{92.13} & 85.82 & 92.86 & 23.25 & 29.16 & 69.58 \\
OPSD & 88.83 & 91.82 & \bestfill{86.08} & \secondfill{95.38} & \secondfill{28.67} & \secondfill{34.26} & 71.65 \\
\textbf{MOPDA} & \secondfill{89.01} & \secondfill{91.91} & \secondfill{85.98} & \bestfill{95.49} & \bestfill{30.58} & \bestfill{35.61} & \bestfill{74.01} \\
\bottomrule
\end{tabular*}
\end{table}

\begin{table}[t]
\centering
\setlength{\belowcaptionskip}{9pt}
\caption{
\textbf{Paired judgment changes under Mixed-Trajectory Supervision across IAD benchmarks.}
OPSD and MOPDA predictions are paired by sample.
Corr./Regr. denote wrong-to-correct and correct-to-wrong transitions.
Anom./Norm. denote class-wise contributions to $\Delta$QA.
}
\label{tab:paired-judgment}

\renewcommand{\arraystretch}{1.28}
\setlength{\tabcolsep}{5.5pt}
\small

\begin{tabular}{lccc cc cc}
\toprule
& \multicolumn{3}{c}{QA (\%)}
& \multicolumn{2}{c}{Paired Changes (\%)}
& \multicolumn{2}{c}{Class-wise $\Delta$QA (\%)} \\
\cmidrule(lr){2-4}
\cmidrule(lr){5-6}
\cmidrule(lr){7-8}
Dataset
& OPSD
& MOPDA
& $\Delta \uparrow$
& Corr. $\uparrow$
& Regr. $\downarrow$
& Anom. $\uparrow$
& Norm. $\uparrow$ \\
\midrule

VisA
& 71.65
& 74.01
& \positivefill{+2.36}
& 5.74\%
& 3.38\%
& \positivefill{+5.32\%}
& \negativefill{$-2.96\%$} \\

SDD
& 71.93
& 83.71
& \positivefill{+11.78}
& 13.78\%
& 2.01\%
& \negativefill{$-1.50\%$}
& \positivefill{+13.28\%} \\

WFDD
& 65.09
& 63.74
& \negativefill{$-1.35$}
& 0.90\%
& 2.25\%
& \negativefill{$-0.90\%$}
& \negativefill{$-0.45\%$} \\

KSDD2
& 86.94
& 87.50
& \positivefill{+0.56}
& 1.69\%
& 1.12\%
& \positivefill{+0.28\%}
& \positivefill{+0.28\%} \\

\bottomrule
\end{tabular}
\end{table}

\begingroup
\setlength{\intextsep}{14pt}

\begin{table}[H]
\centering
\setlength{\belowcaptionskip}{9pt}
\caption{\textbf{Language-guided Visual Anchoring ablation on MVTec-AD.}
The two blocks vary the localization interface and reforwarded judgment.
The MOPDA row matches Table~\ref{tab:main}. The gray GT-judgment row is an
unranked oracle. Best non-oracle results are shaded pink and second-best blue.}
\label{tab:anchor}
\fontsize{8.2pt}{10pt}\selectfont
\renewcommand{\arraystretch}{1.25}
\setlength{\tabcolsep}{3pt}
\begin{tabular}{@{}>{\raggedright\arraybackslash}p{0.37\linewidth}*{6}{>{\centering\arraybackslash}p{0.09\linewidth}}@{}}
\toprule
\multirow{2}{*}{Setting}
& \multicolumn{3}{c}{Image-level}
& \multicolumn{3}{c}{Pixel-level} \\
\cmidrule(lr){2-4}\cmidrule(lr){5-7}
& AUROC & AP & F1 & AUROC & AP & F1 \\
\midrule
\rowcolor{tablegroup}\multicolumn{7}{l}{\textit{Localization interface controls}} \\
\midrule
MOPDA w/ Full Response & 91.49 & 95.75 & 91.84 & 87.63 & 33.21 & 38.30 \\
MOPDA w/ Learnable Anchors Only & 90.95 & 95.56 & 91.44 & 87.35 & 32.63 & 37.65 \\
MOPDA w/o Judgment Aggregation & \secondfill{93.46} & \secondfill{96.89} & \secondfill{92.53} & \secondfill{89.74} & \secondfill{38.72} & \secondfill{42.05} \\
MOPDA w/o Language Output & 90.73 & 95.06 & 91.22 & 85.95 & 27.34 & 33.41 \\
\midrule
\rowcolor{tablegroup}\multicolumn{7}{l}{\textit{Judgment intervention controls}} \\
\midrule
MOPDA w/ Incorrect Judgment & 85.60 & 93.41 & 88.90 & 87.61 & 32.05 & 36.93 \\
\textbf{MOPDA} & \bestfill{93.76} & \bestfill{96.99} & \bestfill{92.58} & \bestfill{90.87} & \bestfill{43.52} & \bestfill{45.44} \\
\midrule
\rowcolor{tableoracle}MOPDA w/ GT Judgment (oracle) & 95.60 & 97.57 & 94.05 & 92.95 & 45.45 & 47.18 \\
\bottomrule
\end{tabular}
\end{table}
\endgroup

\section{Conclusion}
\label{sec:conclusion}

We presented \method, a mixed-trajectory on-policy distillation framework for
LVLM-based industrial anomaly detection. \method combines
\textbf{Mixed-Trajectory Supervision} for judgment learning with
\textbf{Language-guided Visual Anchoring} for dense anomaly localization.
Experiments on five IAD benchmarks demonstrate improved judgment and localization
performance over the evaluated LVLM-based baselines. More importantly, our analysis shows that language judgments provide effective semantic guidance for dense anomaly localization, while dense visual features
remain the primary evidence for pixel-level anomaly scoring.
This provides a flexible connection between language understanding and dense
visual perception without making localization directly determined by the
language output.
\bibliography{references}

\begin{thebibliography}{32}
\providecommand{\natexlab}[1]{#1}
\providecommand{\url}[1]{\texttt{#1}}
\expandafter\ifx\csname urlstyle\endcsname\relax
  \providecommand{\doi}[1]{doi: #1}\else
  \providecommand{\doi}{doi: \begingroup \urlstyle{rm}\Url}\fi

\bibitem[Agarwal et~al.(2024)Agarwal, Vieillard, Zhou, Stanczyk, Ramos~Garea, Geist, and Bachem]{gkd}
Rishabh Agarwal, Nino Vieillard, Yongchao Zhou, Piotr Stanczyk, Sabela Ramos~Garea, Matthieu Geist, and Olivier Bachem.
\newblock On-policy distillation of language models: Learning from self-generated mistakes.
\newblock In B.~Kim, Y.~Yue, S.~Chaudhuri, K.~Fragkiadaki, M.~Khan, and Y.~Sun (eds.), \emph{International Conference on Learning Representations}, volume 2024, pp.\  21246--21263, 2024.

\bibitem[Bai et~al.(2025)Bai, Cai, Chen, Chen, Chen, Cheng, Deng, Ding, Gao, Ge, Ge, Guo, Huang, Huang, Huang, Hui, Jiang, Li, Li, Li, Li, Lin, Lin, Liu, Liu, Liu, Liu, Liu, Liu, Lu, Luo, Lv, Men, Meng, Ren, Ren, Song, Sun, Tang, Tu, Wan, Wang, Wang, Wang, Wang, Xie, Xu, Xu, Xu, Yang, Yang, Yang, Yang, Yu, Zhang, Zhang, Zhang, Zheng, Zhong, Zhou, Zhou, Zhou, Zhu, and Zhu]{qwen3vl}
Shuai Bai, Yuxuan Cai, Ruizhe Chen, Keqin Chen, Xionghui Chen, Zesen Cheng, Lianghao Deng, Wei Ding, Chang Gao, Chunjiang Ge, Wenbin Ge, Zhifang Guo, Qidong Huang, Jie Huang, Fei Huang, Binyuan Hui, Shutong Jiang, Zhaohai Li, Mingsheng Li, Mei Li, Kaixin Li, Zicheng Lin, Junyang Lin, Xuejing Liu, Jiawei Liu, Chenglong Liu, Yang Liu, Dayiheng Liu, Shixuan Liu, Dunjie Lu, Ruilin Luo, Chenxu Lv, Rui Men, Lingchen Meng, Xuancheng Ren, Xingzhang Ren, Sibo Song, Yuchong Sun, Jun Tang, Jianhong Tu, Jianqiang Wan, Peng Wang, Pengfei Wang, Qiuyue Wang, Yuxuan Wang, Tianbao Xie, Yiheng Xu, Haiyang Xu, Jin Xu, Zhibo Yang, Mingkun Yang, Jianxin Yang, An~Yang, Bowen Yu, Fei Zhang, Hang Zhang, Xi~Zhang, Bo~Zheng, Humen Zhong, Jingren Zhou, Fan Zhou, Jing Zhou, Yuanzhi Zhu, and Ke~Zhu.
\newblock Qwen3-vl technical report, 2025.

\bibitem[Bergmann et~al.(2019)Bergmann, Fauser, Sattlegger, and Steger]{mvtecad}
Paul Bergmann, Michael Fauser, David Sattlegger, and Carsten Steger.
\newblock Mvtec ad --- a comprehensive real-world dataset for unsupervised anomaly detection.
\newblock In \emph{2019 IEEE/CVF Conference on Computer Vision and Pattern Recognition (CVPR)}, pp.\  9584--9592. IEEE, June 2019.

\bibitem[Bo{\v{z}}i{\v{c}} et~al.(2021)Bo{\v{z}}i{\v{c}}, Tabernik, and Sko{\v{c}}aj]{ksdd2}
Jakob Bo{\v{z}}i{\v{c}}, Domen Tabernik, and Danijel Sko{\v{c}}aj.
\newblock Mixed supervision for surface-defect detection: From weakly to fully supervised learning.
\newblock \emph{Computers in Industry}, 129:\penalty0 103459, 2021.

\bibitem[Cao et~al.(2024)Cao, Zhang, Frittoli, Cheng, Shen, and Boracchi]{adaclip}
Yunkang Cao, Jiangning Zhang, Luca Frittoli, Yuqi Cheng, Weiming Shen, and Giacomo Boracchi.
\newblock Adaclip: Adapting {CLIP} with hybrid learnable prompts for zero-shot anomaly detection.
\newblock In Ales Leonardis, Elisa Ricci, Stefan Roth, Olga Russakovsky, Torsten Sattler, and G{\"{u}}l Varol (eds.), \emph{Computer Vision - {ECCV} 2024 - 18th European Conference, Milan, Italy, September 29-October 4, 2024, Proceedings, Part {XXXV}}, volume 15093 of \emph{Lecture Notes in Computer Science}, pp.\  55--72. Springer, 2024.

\bibitem[Chao et~al.(2025)Chao, Liu, Tang, and Wu]{anomalyr1}
Yuhao Chao, Jie Liu, Jie Tang, and Gangshan Wu.
\newblock Anomalyr1: A grpo-based end-to-end mllm for industrial anomaly detection, 2025.

\bibitem[Chen et~al.(2024)Chen, Luo, Lv, and Zhang]{wfdd}
Qiyu Chen, Huiyuan Luo, Chengkan Lv, and Zhengtao Zhang.
\newblock A unified anomaly synthesis strategy with gradient ascent for industrial anomaly detection and localization.
\newblock In Ales Leonardis, Elisa Ricci, Stefan Roth, Olga Russakovsky, Torsten Sattler, and G{\"{u}}l Varol (eds.), \emph{Computer Vision - {ECCV} 2024 - 18th European Conference, Milan, Italy, September 29-October 4, 2024, Proceedings, Part {LXVII}}, volume 15125 of \emph{Lecture Notes in Computer Science}, pp.\  37--54. Springer, 2024.

\bibitem[Cheng et~al.(2025)Cheng, Xu, Wang, Ma, He, Zhang, Cai, Zhen, Jia, Wan, Xia, and Zhao]{anomalycot}
Jiaxi Cheng, Yuliang Xu, Shoupeng Wang, Tao Ma, Yuchen He, Jinghe Zhang, Sihang Cai, Jiawei Zhen, Jingyi Jia, Yao Wan, Yan Xia, and Zhou Zhao.
\newblock {AnomalyCoT}: A multi-scenario chain-of-thought dataset for multimodal large language models.
\newblock In \emph{Advances in Neural Information Processing Systems}, volume~38, 2025.
\newblock \doi{10.52202/085713-2332}.

\bibitem[Deng et~al.(2026)Deng, Luo, Zhai, Guo, Cao, and Kang]{vmad}
Huilin Deng, Hongchen Luo, Wei Zhai, Yanming Guo, Yang Cao, and Yu~Kang.
\newblock Vmad: Visual-enhanced multimodal large language model for zero-shot anomaly detection.
\newblock \emph{IEEE Transactions on Automation Science and Engineering}, 23:\penalty0 3607--3618, 2026.

\bibitem[Gu et~al.(2024{\natexlab{a}})Gu, Dong, Wei, and Huang]{minillm}
Yuxian Gu, Li~Dong, Furu Wei, and Minlie Huang.
\newblock Minillm: Knowledge distillation of large language models.
\newblock In B.~Kim, Y.~Yue, S.~Chaudhuri, K.~Fragkiadaki, M.~Khan, and Y.~Sun (eds.), \emph{International Conference on Learning Representations}, volume 2024, pp.\  32694--32717, 2024{\natexlab{a}}.

\bibitem[Gu et~al.(2024{\natexlab{b}})Gu, Zhu, Zhu, Chen, Tang, and Wang]{anomalygpt}
Zhaopeng Gu, Bingke Zhu, Guibo Zhu, Yingying Chen, Ming Tang, and Jinqiao Wang.
\newblock Anomalygpt: Detecting industrial anomalies using large vision-language models.
\newblock \emph{Proceedings of the AAAI Conference on Artificial Intelligence}, 38\penalty0 (3):\penalty0 1932--1940, Mar 2024{\natexlab{b}}.

\bibitem[He et~al.(2026)He, Meng, Meng, and Yang]{adopd}
Jingtai He, Shiyuan Meng, Wenchao Meng, and Qinmin Yang.
\newblock Adopd: Reference-privileged on-policy distillation for mllm-based industrial anomaly detection, 2026.

\bibitem[Hou et~al.(2026)Hou, Li, Liu, Wang, Ruan, Qiu, and Xu]{visualad}
Yanning Hou, Peiyuan Li, Zirui Liu, Yitong Wang, Yanran Ruan, Jianfeng Qiu, and Ke~Xu.
\newblock Visualad: Language-free zero-shot anomaly detection via vision transformer.
\newblock In \emph{Proceedings of the IEEE/CVF Conference on Computer Vision and Pattern Recognition (CVPR)}, pp.\  21346--21356, June 2026.

\bibitem[Jeong et~al.(2023)Jeong, Zou, Kim, Zhang, Ravichandran, and Dabeer]{winclip}
Jongheon Jeong, Yang Zou, Taewan Kim, Dongqing Zhang, Avinash Ravichandran, and Onkar Dabeer.
\newblock Winclip: Zero-/few-shot anomaly classification and segmentation.
\newblock In \emph{Proceedings of the IEEE/CVF Conference on Computer Vision and Pattern Recognition (CVPR)}, pp.\  19606--19616, June 2023.

\bibitem[Jiang et~al.(2025)Jiang, Li, Deng, Liu, Gao, Zhou, Li, Wang, and Zheng]{mmad}
Xi~Jiang, Jian Li, Hanqiu Deng, Yong Liu, Bin-Bin Gao, Yifeng Zhou, Jialin Li, Chengjie Wang, and Feng Zheng.
\newblock {MMAD}: A comprehensive benchmark for multimodal large language models in industrial anomaly detection.
\newblock In \emph{ICLR}, 2025.

\bibitem[Jin et~al.(2026)Jin, Feng, Zhang, Wang, Liu, and Wang]{real}
Yizhou Jin, Yuezhu Feng, Jinjin Zhang, Peng Wang, Qingjie Liu, and Yunhong Wang.
\newblock Reasoning-driven anomaly detection and localization with image-level supervision.
\newblock In \emph{Proceedings of the IEEE/CVF Conference on Computer Vision and Pattern Recognition (CVPR)}, pp.\  43061--43071, June 2026.

\bibitem[Li et~al.(2024)Li, Huang, Xue, and Zhou]{musc}
Xurui Li, Ziming Huang, Feng Xue, and Yu~Zhou.
\newblock Musc: Zero-shot industrial anomaly classification and segmentation with mutual scoring of the unlabeled images.
\newblock In B.~Kim, Y.~Yue, S.~Chaudhuri, K.~Fragkiadaki, M.~Khan, and Y.~Sun (eds.), \emph{International Conference on Learning Representations}, volume 2024, pp.\  2172--2197, 2024.

\bibitem[Li et~al.(2026{\natexlab{a}})Li, Cao, Liu, Xiong, Dong, and Huang]{iadr1}
Yanhui Li, Yunkang Cao, Chengliang Liu, Yuan Xiong, Xinghui Dong, and Chao Huang.
\newblock Iad-r1: Reinforcing consistent reasoning in industrial anomaly detection.
\newblock \emph{Proceedings of the AAAI Conference on Artificial Intelligence}, 40\penalty0 (8):\penalty0 6583--6591, Mar 2026{\natexlab{a}}.

\bibitem[Li et~al.(2026{\natexlab{b}})Li, Wang, Yuan, Liu, Zhao, Guo, Xu, Shi, and Zuo]{myriad}
Yuanze Li, Haolin Wang, Shihao Yuan, Ming Liu, Debin Zhao, Yiwen Guo, Chen Xu, Guangming Shi, and Wangmeng Zuo.
\newblock Myriad: A large multimodal model applying vision experts for industrial anomaly detection.
\newblock \emph{Science China Information Sciences}, 69\penalty0 (9), Aug 2026{\natexlab{b}}.

\bibitem[Lin et~al.(2017)Lin, Goyal, Girshick, He, and Dollar]{focal}
Tsung-Yi Lin, Priya Goyal, Ross Girshick, Kaiming He, and Piotr Dollar.
\newblock Focal loss for dense object detection.
\newblock In \emph{2017 IEEE International Conference on Computer Vision (ICCV)}, pp.\  2999--3007. IEEE, Oct 2017.

\bibitem[Ma et~al.(2026)Ma, Cheng, Li, Xu, Tan, Shao, Wang, Liu, Gao, Wang, et~al.]{ma2026hug}
Fei Ma, Zebang Cheng, Minghui Li, Hongbo Xu, Yuyong Tan, Yihua Shao, Hanling Wang, Zhou Liu, Yuqing Gao, Dong Wang, et~al.
\newblock Hug-vis: A multimodal benchmark for human-centered understanding and generation in visual intelligence.
\newblock \emph{arXiv preprint arXiv:2608.26517}, 2026.

\bibitem[Milletari et~al.(2016)Milletari, Navab, and Ahmadi]{dice}
Fausto Milletari, Nassir Navab, and Seyed-Ahmad Ahmadi.
\newblock V-net: Fully convolutional neural networks for volumetric medical image segmentation.
\newblock In \emph{2016 Fourth International Conference on 3D Vision (3DV)}, pp.\  565--571. IEEE, Oct 2016.

\bibitem[Qu et~al.(2026)Qu, Tao, Bao, Wang, Qu, Zhang, and Wang]{agvas}
Zhen Qu, Xian Tao, Xiaoyi Bao, Dingrong Wang, ShiChen Qu, Zhengtao Zhang, and Xingang Wang.
\newblock Ag-vas: Anchor-guided zero-shot visual anomaly segmentation with large multimodal models.
\newblock In \emph{Proceedings of the IEEE/CVF Conference on Computer Vision and Pattern Recognition (CVPR)}, pp.\  14126--14136, June 2026.

\bibitem[Shao et~al.(2024)Shao, Wang, Zhu, Xu, Song, Bi, Zhang, Zhang, Li, Wu, and Guo]{deepseekmath}
Zhihong Shao, Peiyi Wang, Qihao Zhu, Runxin Xu, Junxiao Song, Xiao Bi, Haowei Zhang, Mingchuan Zhang, Y.~K. Li, Y.~Wu, and Daya Guo.
\newblock Deepseekmath: Pushing the limits of mathematical reasoning in open language models, 2024.

\bibitem[Tabernik et~al.(2019)Tabernik, {\v{S}}ela, Skvar{\v{c}}, and Sko{\v{c}}aj]{kolektorsdd}
Domen Tabernik, Samo {\v{S}}ela, Jure Skvar{\v{c}}, and Danijel Sko{\v{c}}aj.
\newblock Segmentation-based deep-learning approach for surface-defect detection.
\newblock \emph{Journal of Intelligent Manufacturing}, 31\penalty0 (3):\penalty0 759--776, May 2019.

\bibitem[Tarvainen \& Valpola(2017)Tarvainen and Valpola]{tarvainen2017mean}
Antti Tarvainen and Harri Valpola.
\newblock Mean teachers are better role models: Weight-averaged consistency targets improve semi-supervised deep learning results.
\newblock In \emph{Advances in Neural Information Processing Systems 30: Annual Conference on Neural Information Processing Systems 2017}, pp.\  1195--1204, 2017.

\bibitem[Xu et~al.(2025)Xu, Lo, Safaei, Patel, and Dwivedi]{anomalyov}
Jiacong Xu, Shao-Yuan Lo, Bardia Safaei, Vishal~M. Patel, and Isht Dwivedi.
\newblock Towards zero-shot anomaly detection and reasoning with multimodal large language models.
\newblock In \emph{Proceedings of the IEEE/CVF Conference on Computer Vision and Pattern Recognition (CVPR)}, pp.\  20370--20382, June 2025.

\bibitem[Zhang et~al.(2024)Zhang, Ding, Ban, and Dai]{goodsad}
Jian Zhang, Runwei Ding, Miaoju Ban, and Linhui Dai.
\newblock Pku-goodsad: A supermarket goods dataset for unsupervised anomaly detection and segmentation.
\newblock \emph{IEEE Robotics and Automation Letters}, 9\penalty0 (3):\penalty0 2008--2015, Mar 2024.

\bibitem[Zhao et~al.(2025)Zhao, Lin, Han, Zhao, and Wei]{omniad}
Shifang Zhao, Yiheng Lin, Lu~Han, Yao Zhao, and Yunchao Wei.
\newblock Omniad: Detect and understand industrial anomaly via multimodal reasoning, 2025.

\bibitem[Zhao et~al.(2026)Zhao, Xie, Liu, Huang, Pang, Chen, and Grover]{opsd}
Siyan Zhao, Zhihui Xie, Mengchen Liu, Jing Huang, Guan Pang, Feiyu Chen, and Aditya Grover.
\newblock Self-distilled reasoner: On-policy self-distillation for large language models, 2026.

\bibitem[Zhou et~al.(2024)Zhou, Pang, Tian, He, and Chen]{anomalyclip}
Qihang Zhou, Guansong Pang, Yu~Tian, Shibo He, and Jiming Chen.
\newblock Anomalyclip: Object-agnostic prompt learning for zero-shot anomaly detection.
\newblock In B.~Kim, Y.~Yue, S.~Chaudhuri, K.~Fragkiadaki, M.~Khan, and Y.~Sun (eds.), \emph{International Conference on Learning Representations}, volume 2024, pp.\  49705--49737, 2024.

\bibitem[Zou et~al.(2022)Zou, Jeong, Pemula, Zhang, and Dabeer]{visa}
Yang Zou, Jongheon Jeong, Latha Pemula, Dongqing Zhang, and Onkar Dabeer.
\newblock \emph{SPot-the-Difference Self-supervised Pre-training for Anomaly Detection and Segmentation}, pp.\  392--408.
\newblock Springer Nature Switzerland, 2022.

\end{thebibliography}
\bibliographystyle{iclr2027_conference}

\clearpage
\appendix
\raggedbottom
\setlength{\intextsep}{7pt}
\setlength{\abovecaptionskip}{7pt}
\definecolor{tablecyan}{HTML}{EAF8F8}
\definecolor{tableorange}{HTML}{FFF1E3}
% Appendix title and first figure.
\emergencystretch=3em
\noindent{\LARGE\sc MOPDA: Mixed-Trajectory On-Policy Distillation for
Language-Guided Industrial Anomaly Detection\par}
{\Large\sc Supplementary Material\par}

\begin{figure}[H]
\centering
\includegraphics[width=\linewidth]{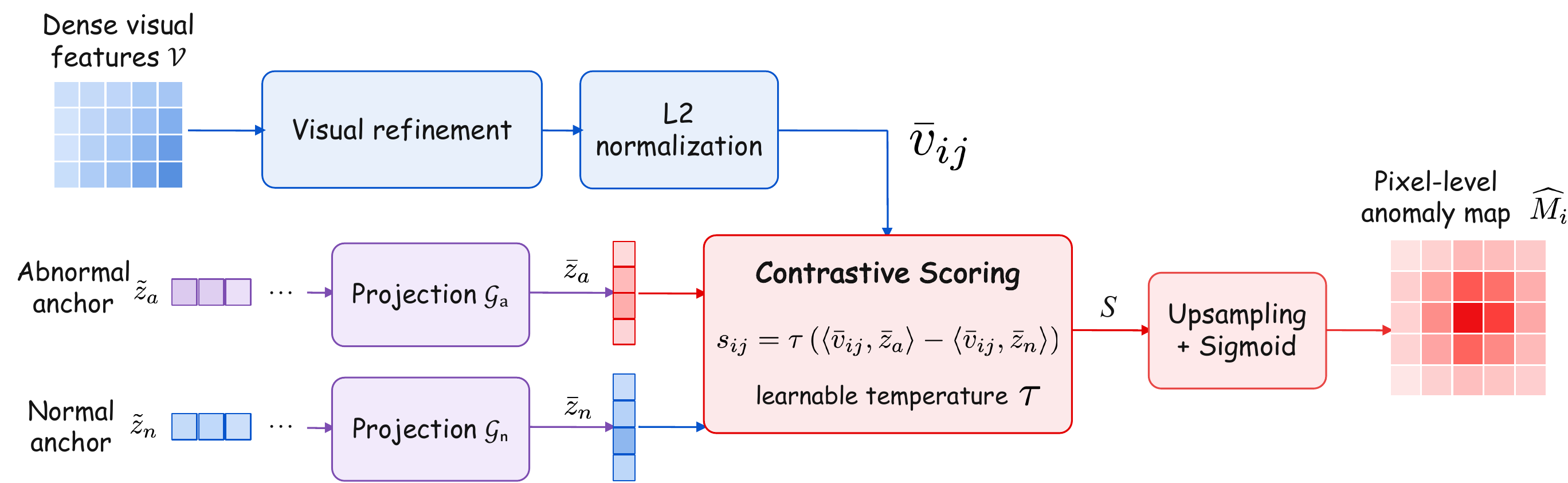}
\caption{\textbf{Architecture of the contrastive heatmap head.}
Dense visual features and judgment-conditioned abnormal and normal anchors
are mapped to a shared 256-dimensional space. Their scaled cosine-similarity
difference produces anomaly logits, which are upsampled and passed through
a sigmoid to form the pixel-level anomaly map. Image-level aggregation is
described below.}
\label{fig:app-contrastive-head}
\end{figure}

% Appendix content for main.tex.

\section{Implementation and Training Details}
\label{app:training-details}

\subsection{Language-guided Visual Anchoring Details}
\label{app:anchoring-details}

\paragraph{Judgment extraction.}
For a selected training response or the student's inference response $\hat y_i$,
$\hat a_i=\operatorname{Trim}_{\mathrm{ans}}(\hat y_i)$ extracts the final span
enclosed by \texttt{<answer></answer>}. This span is both the language-side
judgment and the condition supplied to the judgment reforward.

\paragraph{Backbone and head inputs.}
Figure~\ref{fig:app-contrastive-head} shows the localization head used with
Qwen3-VL-8B-Instruct. We reshape the query image's post-merger visual tokens
into a single 4096-channel spatial grid $V_i$. The judgment reforward yields
4096-dimensional contextualized hidden states $\tilde z_a$ and $\tilde z_n$
at the abnormal and normal anchor tokens. These are the head's language inputs.

\paragraph{Visual refinement and anchor projection.}
The visual grid first passes through a learned residual channel adapter. It
applies channel-wise LayerNorm and two $1\times1$ convolutions with a
$4096\!\rightarrow\!1024\!\rightarrow\!4096$ bottleneck and GELU; its output
is added to the input with a learnable scale initialized to zero. Two
$3\times3$ Conv--GELU layers then map the grid to 256 channels. Separately,
$g_a$ and $g_n$ project the two anchor states to 256 dimensions using
Linear--GELU--Linear--LayerNorm. Denoting the refined visual grid by $h(V_i)$,
the representations used for scoring are
\begin{equation}
\bar v_{ij}=\frac{h(V_i)_j}{\|h(V_i)_j\|_2},
\qquad
\bar z_a=\frac{g_a(\tilde z_a)}{\|g_a(\tilde z_a)\|_2},
\qquad
\bar z_n=\frac{g_n(\tilde z_n)}{\|g_n(\tilde z_n)\|_2}.
\label{eq:anchor-normalization}
\end{equation}
Each patch is scored by the abnormal-minus-normal cosine-similarity
difference, multiplied by the learnable temperature
$\tau=\min(\exp(\alpha),100)$, with learnable $\alpha$.

\paragraph{Inference outputs.}
At inference, the logits are bilinearly upsampled to the
evaluation resolution and passed through Sigmoid to obtain the anomaly map.
The image-level anomaly score is the mean of the largest $1\%$ of values in
this map. For Gaussian-smoothed evaluation, smoothing is applied to the
probability map before computing the image score and localization metrics.

\newpage
\subsection{Distillation Implementation Details}
\label{app:distillation-details}

\paragraph{EMA teacher.}
The teacher has the same architecture and vocabulary as the student. Its
parameters are maintained by an exponential moving average (EMA) of the student
parameters~\citep{tarvainen2017mean},
\begin{equation}
\bar\theta^{(k)}=\mu\bar\theta^{(k-1)}+(1-\mu)\theta^{(k)},
\qquad \mu\in(0,1).
\label{eq:ema-teacher}
\end{equation}
We set $\mu=0.999$ in the reported experiments.

\paragraph{Generalized Jensen--Shannon divergence.}
For either trajectory source, the student and teacher evaluate the same
sampled response prefixes under their respective contexts. Their next-token
distributions at position $t$ are
\begin{equation}
P_{S,t}^{(i)}
=p_\theta(\cdot\mid c_i^S,\hat y_{i,<t}),
\qquad
P_{T,t}^{(i)}
=\operatorname{stopgrad}\!\left(
p_{\bar\theta}(\cdot\mid c_i^T,\hat y_{i,<t})
\right).
\label{eq:distillation-distributions}
\end{equation}
We use generalized Jensen--Shannon divergence with $\beta\in(0,1)$,
\begin{equation}
\begin{aligned}
m_{i,t}^\beta&=(1-\beta)P_{S,t}^{(i)}+\beta P_{T,t}^{(i)},\\
D_{\mathrm{JSD}}^\beta(P_{S,t}^{(i)}\|P_{T,t}^{(i)})
&=\beta\,\mathrm{KL}(P_{T,t}^{(i)}\|m_{i,t}^\beta)
 +(1-\beta)\,\mathrm{KL}(P_{S,t}^{(i)}\|m_{i,t}^\beta).
\end{aligned}
\label{eq:distillation-jsd}
\end{equation}
With $\beta=0.5$, this is the symmetric JSD. Gradients flow only through the
student distributions, not through the teacher distributions or sampled tokens.

\paragraph{Teacher-sequence validity and exact loss.}
Teacher-generated responses are checked for the required format and agreement
of their final judgment with the source label $a_i$. We set $g_i=1$ for
student-generated trajectories and
$g_i=\operatorname{Valid}(\hat y_i,a_i)\in\{0,1\}$ for teacher-generated
trajectories. This gate applies only to distillation. The exact minibatch loss
and token-count normalization are
\begin{equation}
\mathcal L_{\mathrm{dist}}
=\frac{1}{N_{\mathcal B}}
\sum_{i\in\mathcal B}g_i\sum_{t=1}^{|\hat y_i|}
D_{\mathrm{JSD}}^\beta(P_{S,t}^{(i)}\|P_{T,t}^{(i)}),
\qquad
N_{\mathcal B}=\sum_{i\in\mathcal B}|\hat y_i|.
\label{eq:exact-distillation-loss}
\end{equation}
Invalid teacher responses contribute zero to the numerator. The denominator
still counts all sampled tokens in the minibatch, as in the main-text objective.

\subsection{Training Objective and Optimization Details}
\label{app:training-objective}

\paragraph{Localization supervision.}
During source-domain training, the mean of the largest $1\%$ of low-resolution
anomaly logits is aligned with the image label $a_i$ by BCE, yielding
$\mathcal L_{\mathrm{img}}$. This Top-1\% pooling is omitted from
Figure~\ref{fig:app-contrastive-head}. Pixel-level supervision compares the
anomaly response with mask $M_i$ using BCE, focal loss~\citep{focal}, and
Dice loss~\citep{dice}, denoted by
$\mathcal L_{\mathrm{pix}}$, $\mathcal L_{\mathrm{foc}}$, and
$\mathcal L_{\mathrm{dice}}$. Following~\citep{visualad}, the contextualized
abnormal and normal anchors are separated with
\begin{equation}
\mathcal L_{\mathrm{con}}
=\max(\langle \bar z_a,\bar z_n\rangle+m,0).
\label{eq:anchor-separation-loss}
\end{equation}
The complete localization objective is
\begin{equation}
\mathcal L_{\mathrm{map}}
=\lambda_{\mathrm{img}}\mathcal L_{\mathrm{img}}
+\lambda_{\mathrm{pix}}\mathcal L_{\mathrm{pix}}
+\lambda_{\mathrm{foc}}\mathcal L_{\mathrm{foc}}
+\lambda_{\mathrm{dice}}\mathcal L_{\mathrm{dice}}
+\lambda_{\mathrm{con}}\mathcal L_{\mathrm{con}}.
\label{eq:localization-loss}
\end{equation}
The reported Qwen3-VL-8B-Instruct main experiments use
$\lambda_{\mathrm{img}}=2.0$,
$\lambda_{\mathrm{pix}}=0.5$, and
$\lambda_{\mathrm{foc}}=\lambda_{\mathrm{dice}}=\lambda_{\mathrm{con}}=1.0$.
The checkpoints at training steps 1200 and 2000 share these weights.

\paragraph{Outcome reward and branch updates.}
Student-generated trajectories receive a lightweight reward for response
format and final-judgment correctness:
\begin{equation}
R_i=R_{\mathrm{fmt}}(\hat y_i)+R_{\mathrm{ans}}(\hat a_i,a_i).
\label{eq:outcome-reward}
\end{equation}
The two binary components are
\begin{equation}
\begin{aligned}
R_{\mathrm{fmt}}(\hat y_i)
&=\mathbf{1}\!\left[
\operatorname{Has}_{\mathrm{think}}(\hat y_i)
\land\operatorname{Has}_{\mathrm{answer}}(\hat y_i)\right],\\
R_{\mathrm{ans}}(\hat a_i,a_i)
&=\mathbf{1}\!\left[\operatorname{Label}(\hat a_i)=a_i\right].
\end{aligned}
\label{eq:reward-components}
\end{equation}
Each $\operatorname{Has}$ check requires paired opening and closing tags.
$\operatorname{Label}$ maps a parsed \texttt{A. anomaly} or \texttt{B. normal}
answer to its binary class and is invalid if parsing fails. The corresponding
policy-gradient term $\mathcal L_{\mathrm{reward}}$ provides an auxiliary
outcome signal. For the teacher-generated Seq-KD branch,
$\mathcal L_{\mathrm{reward}}=0$. The joint objective combines
$\mathcal L_{\mathrm{dist}}$,
$\lambda_R\mathcal L_{\mathrm{reward}}$, and
$\lambda_H\mathcal L_{\mathrm{map}}$.
The reported experiments use $\lambda_R=0.1$ and $\lambda_H=1.0$.
The latter is \texttt{lambda\_seg} in the training configuration and is
separate from the five component weights in Eq.~\eqref{eq:localization-loss}.

% Shared training algorithm for the full and standalone appendices.
% Requires algorithm, algpseudocode, amsmath, and amssymb.
\begin{algorithm}[!t]
\caption{MOPDA Training via Mixed-Trajectory On-Policy Distillation}
\label{alg:mopda-training}
\footnotesize
\begin{algorithmic}[1]
\Require Source data $\mathcal D_{\rm src}=\{(x_i^{\rm img},x_i^{\rm text},a_i,M_i,r_i)\}$, student parameters $\theta$
\Statex \textbf{Hyperparameters:} Student-rollout probability $\lambda$, JSD coefficient $\beta$, EMA decay $\mu$,
\Statex \hspace{\algorithmicindent}reward and localization weights $\lambda_R,\lambda_H$
\State Initialize EMA teacher $\bar\theta\gets\theta$
\For{each minibatch $\mathcal B\subset\mathcal D_{\rm src}$}
    \State Construct student contexts $c_i^S$ and evidence-privileged teacher contexts $c_i^T$ for $i\in\mathcal B$
    \State Sample $u\sim\mathcal U(0,1)$
    \If{$u\leq\lambda$} \Comment{student rollout}
        \State Sample $\hat y_i\sim\pi_\theta(\cdot\mid c_i^S)$, set $g_i\gets1$, and compute reward loss $\mathcal L_{\rm reward}$
    \Else \Comment{teacher rollout (Seq-KD)}
        \State Sample $\hat y_i\sim\pi_{\bar\theta}(\cdot\mid c_i^T)$, set $g_i\gets\operatorname{Valid}(\hat y_i,a_i)$ and $\mathcal L_{\rm reward}\gets0$
    \EndIf
    \State Compute $\mathcal L_{\rm dist}$ using Eq.~(\ref{eq:exact-distillation-loss})
    \State Compute $\mathcal L_{\rm map}$ from source labels, masks, and anchor separation
    \State Form $\mathcal L\gets\mathcal L_{\rm dist}+\lambda_R\mathcal L_{\rm reward}+\lambda_H\mathcal L_{\rm map}$
    \State Update student parameters $\theta$ by minimizing $\mathcal L$
    \State $\bar\theta\gets\mu\bar\theta+(1-\mu)\theta$
\EndFor
\end{algorithmic}
\end{algorithm}

\subsection{Training Configuration and Diagnostics}

\paragraph{Backbone and fine-tuning.}
We use Qwen3-VL-8B-Instruct as the vision-language backbone. The reported
\method model uses full-parameter fine-tuning, including the visual tower.
Both language and dense visual representations adapt to source-domain training.

\paragraph{Controlled comparisons.}
All training-paradigm and localization-interface ablations use the same
backbone, source data, and trainable visual tower. They differ only in the
component being studied, such as replacing Mixed-Trajectory On-Policy
Distillation with supervised fine-tuning, removing the generated judgment from the localization
path, or changing the text used to condition the anchors.

\begin{figure}[H]
\centering
\includegraphics[width=\linewidth]{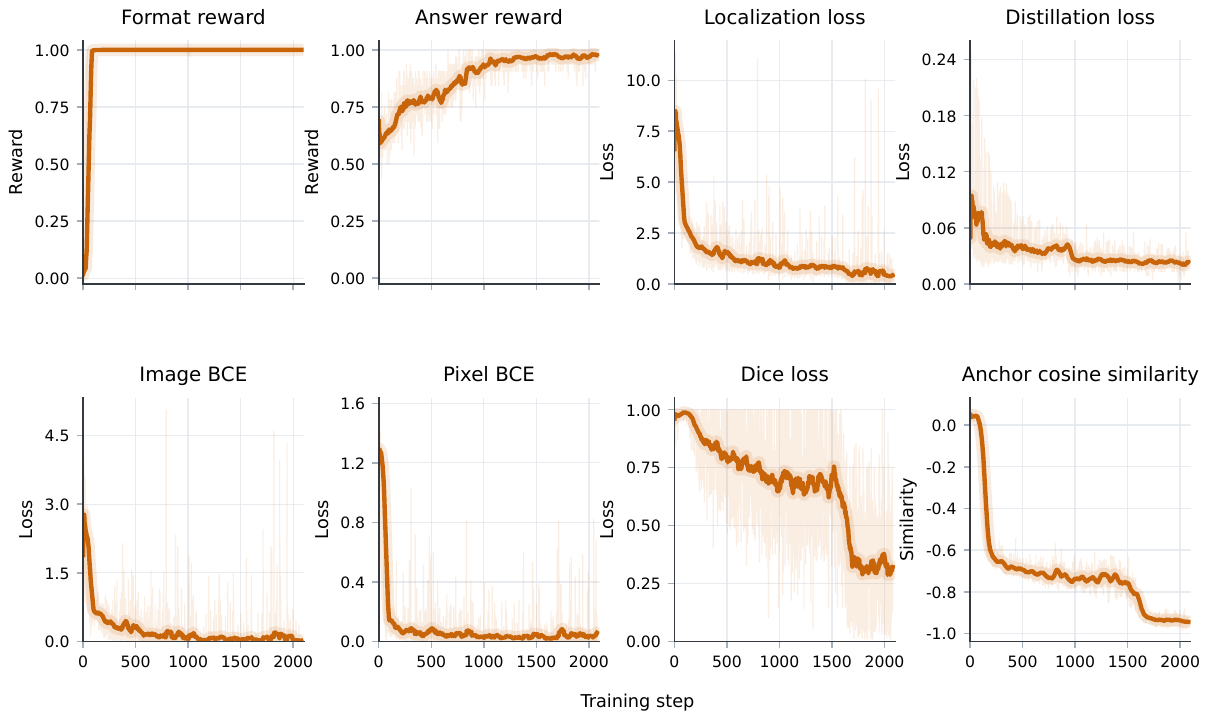}
\caption{\textbf{Training dynamics of MOPDA.}}
\label{fig:app-qwen3-training-dynamics}
\end{figure}

\paragraph{Training dynamics and optimization diagnostics.}
Figure~\ref{fig:app-qwen3-training-dynamics} shows the optimization dynamics of the
language and localization objectives throughout training.
The format reward rapidly saturates, indicating that the model quickly learns
the required response structure, while the answer reward improves progressively
as training proceeds.
Meanwhile, both the localization and distillation losses decrease steadily.
The image- and pixel-level BCE losses also rapidly converge to low values,
whereas the Dice loss continues to decrease in later training, reflecting
progressive refinement of spatial overlap.
Finally, the cosine similarity between the normal and abnormal anchors becomes
increasingly negative, indicating that the anchor-separation objective
progressively encourages more discriminative normal--abnormal representations.
Overall, these trends show stable joint optimization of language judgment,
distillation, and dense anomaly localization.

\section{Dataset Construction and Zero-Shot Protocol}
\label{app:data-construction}

\paragraph{Source-target separation.}
We follow a strict cross-dataset protocol: images from a target benchmark are
not used to train the model evaluated on that benchmark. The target-domain
characteristics and source-target assignments are summarized in
Table~\ref{tab:app-protocol}. GoodsAD is used as an additional source dataset,
while every reported target remains held out.

\begin{table}[h]
\centering
\caption{\textbf{Cross-dataset source-target protocol.} The table summarizes
the visual characteristics of each held-out target benchmark and the
source-domain data used to train the corresponding model.}
\label{tab:app-protocol}
\small
\renewcommand{\arraystretch}{1.15}
\begin{tabular}{@{}l
  >{\raggedright\arraybackslash}p{0.31\linewidth}
  l@{}}
\toprule
\rowcolor{tablegroup}
Target benchmark & Target characteristics & Source-domain training data \\
\midrule
MVTec-AD & Objects and textures & VisA + GoodsAD \\
VisA & Industrial objects & MVTec-AD + GoodsAD \\
SDD/KolektorSDD & Steel-surface defects & MVTec-AD + GoodsAD \\
WFDD & Fabric defects & MVTec-AD + GoodsAD \\
KSDD2 & Metal-surface defects & MVTec-AD + GoodsAD \\
\bottomrule
\end{tabular}
\end{table}

\paragraph{Source-domain supervision.}
Source samples include normal and anomalous images. Pixel masks supervise
dense localization, image-level labels supervise binary anomaly discrimination,
and reference answers construct the source-domain teacher context.

\paragraph{Reference evidence.}
The training-only defect-analysis references $r_i$ were sourced from the open
AnomalyCoT~\citep{anomalycot} and MMAD~\citep{mmad} datasets. They were
supplied only to the evidence-privileged teacher during source-domain training.

\paragraph{Target-domain evaluation.}
For every target benchmark, the model is evaluated directly without
target-domain fine-tuning, target-specific prompt optimization, or access to
target labels and masks. This protocol evaluates cross-domain transfer of the
learned connection between language judgment and visual localization.

\section{Prompt Design}
\label{app:prompt-design}

\begin{figure}[H]
\centering
\includegraphics[width=0.96\linewidth]{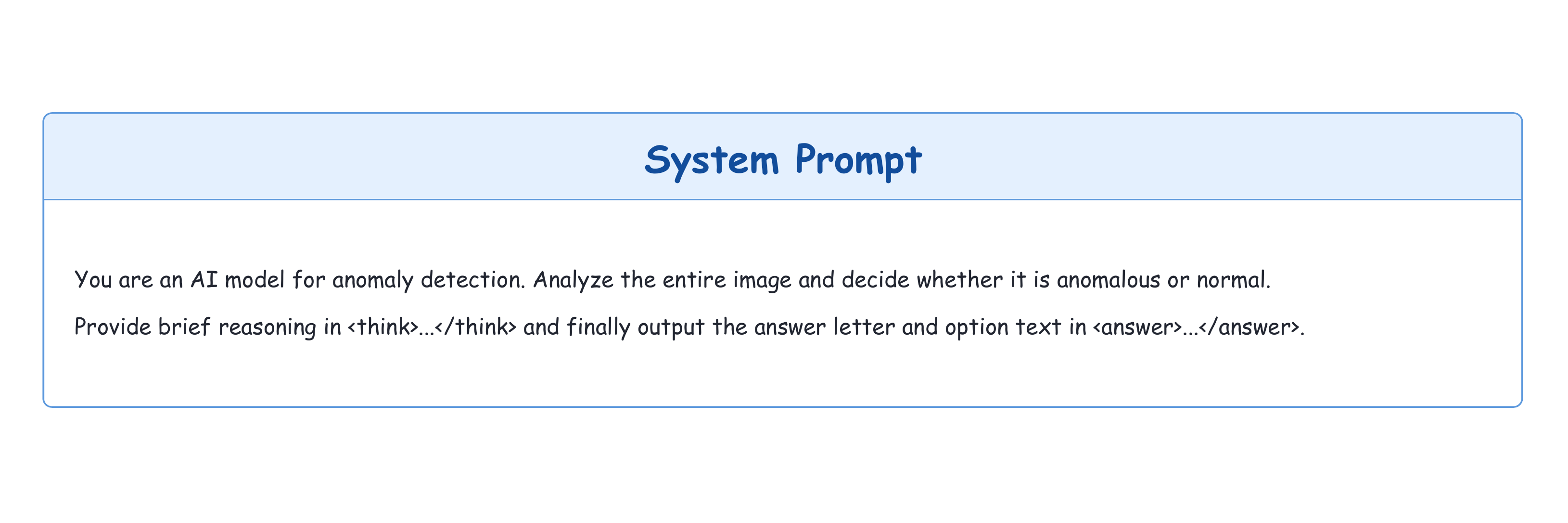}
\caption{\textbf{System Prompt.} The shared anomaly-detection role and
structured response format used by both branches.}
\label{fig:app-prompt-system}
\end{figure}

\begin{figure}[H]
\centering
\includegraphics[width=0.96\linewidth]{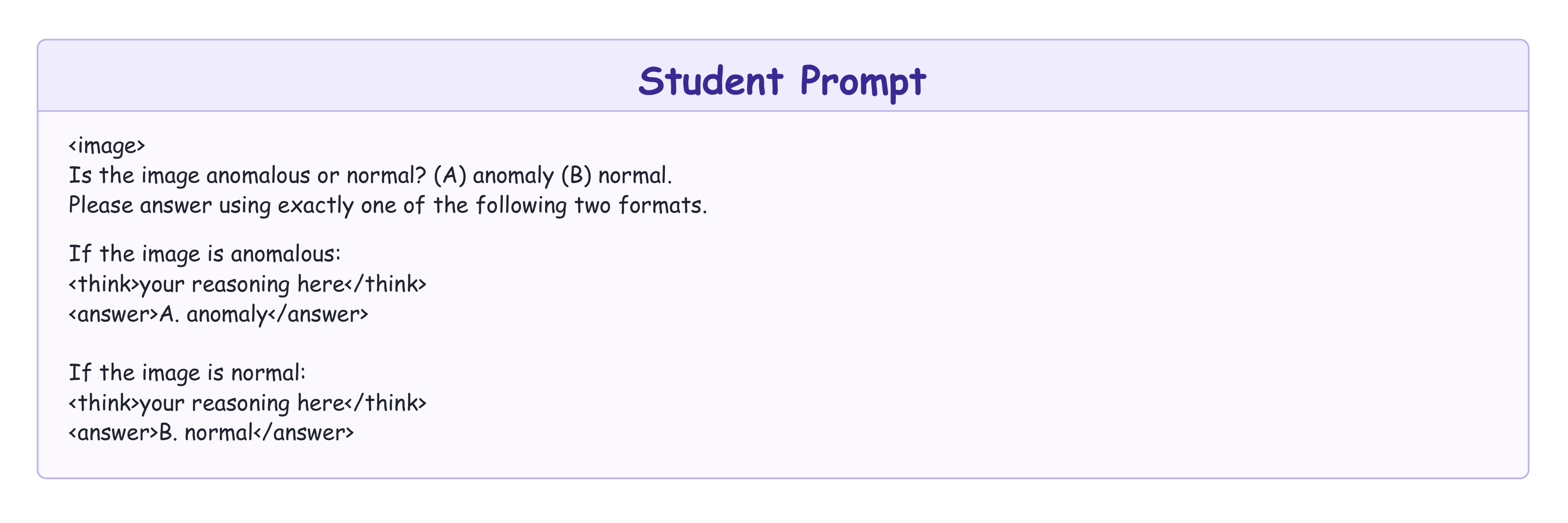}
\caption{\textbf{Student Prompt.} The deployment-compatible query contains only
the image and the binary anomaly question.}
\label{fig:app-prompt-student}
\end{figure}

\begin{figure}[H]
\centering
\includegraphics[width=0.96\linewidth]{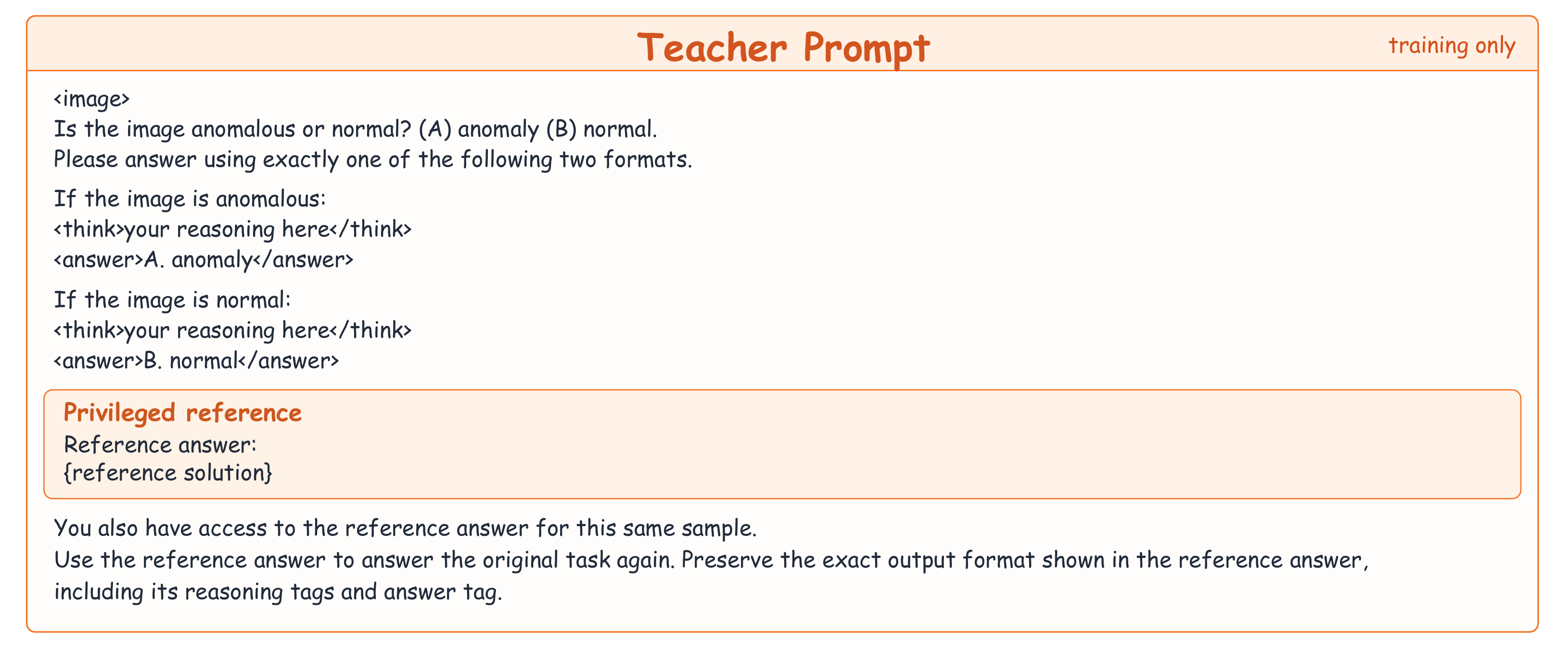}
\caption{\textbf{Teacher Prompt.} A verified source-domain reference answer is
appended only for training-time teacher supervision.}
\label{fig:app-prompt-teacher}
\end{figure}

\paragraph{Prompt organization.}
Figures~\ref{fig:app-prompt-system}--\ref{fig:app-prompt-teacher} make the
information boundary between the student and teacher explicit. The shared
system prompt defines the anomaly-detection role and structured output. The
student prompt contains only the deployment-compatible image and question,
whereas the teacher prompt preserves the same task but appends a verified
reference answer during training. Thus, privileged evidence is confined to
teacher supervision and never enters the deployed input.

\ifdefined\opdiadincludeappendix\newpage\fi
\paragraph{Connection to localization.}
The student's final answer span supplies the compact semantic condition for
visual anchoring. The complete response contributes to language learning, while
localization uses the extracted judgment.

\section{Evaluation Details}
\label{app:evaluation-details}

\paragraph{Evaluation outputs.}
Each target image produces two complementary predictions: a binary language
judgment and a dense anomaly map. The binary judgment is evaluated with
accuracy. The anomaly map is evaluated at both image and pixel levels using
AUROC, average precision, and F1.

\paragraph{Full-resolution localization.}
Pixel-level metrics are computed against the original target masks at their
native evaluation resolution. Image-level anomaly scores are obtained by
aggregating the most anomalous spatial responses in the predicted map.

\paragraph{No target supervision.}
Evaluation uses only the image and the common binary anomaly question. Target
labels and masks are read only by the metric computation after prediction; they
are never exposed to response generation, judgment conditioning, or heatmap
prediction.

\section{More Case Studies}
\label{app:case-studies}

\subsection{Qualitative Localization Comparisons}
\label{app:qualitative-localization}

\begin{figure}[H]
    \centering
    \includegraphics[width=0.76\textwidth]{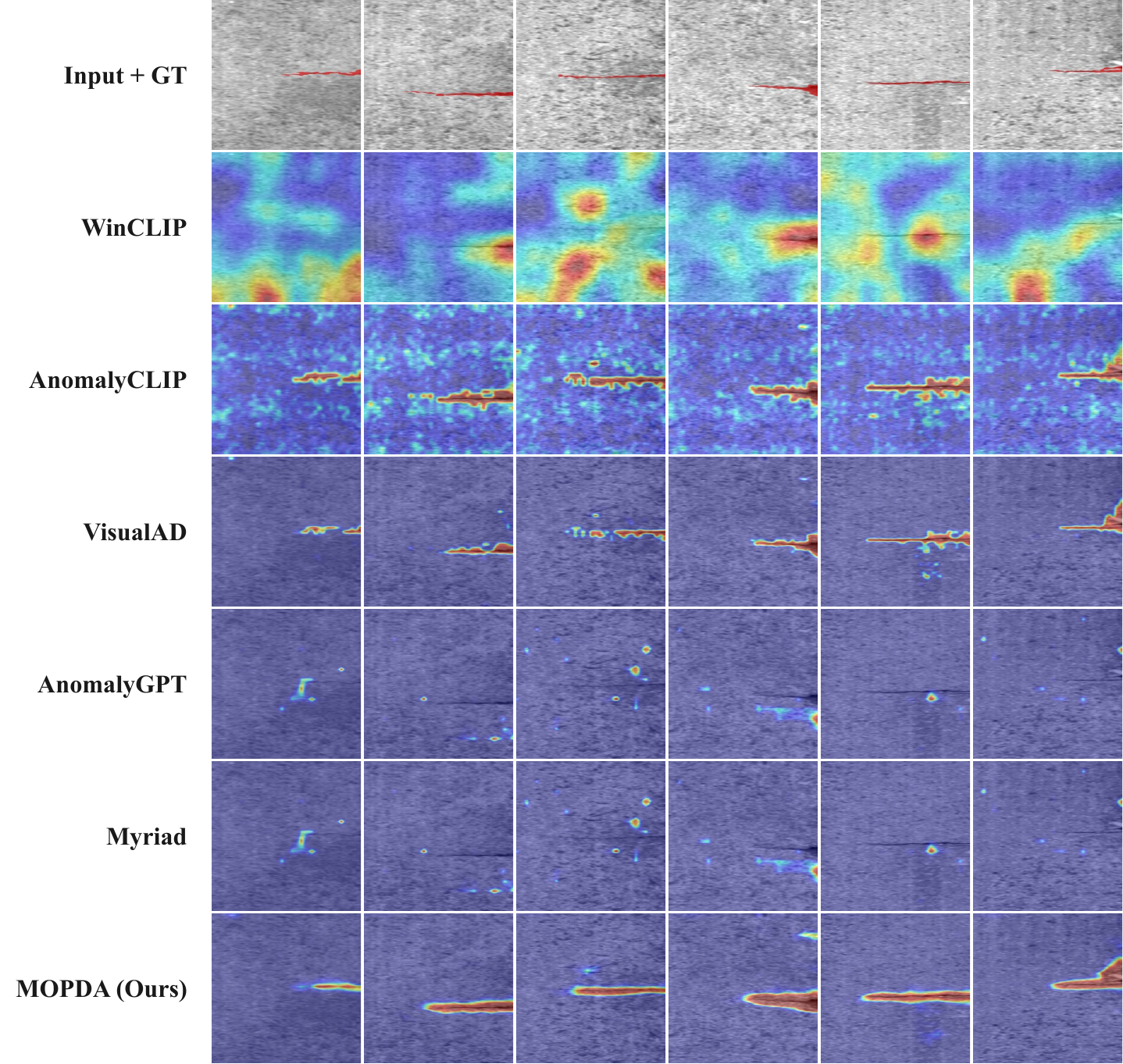}
    \caption{\textbf{Qualitative comparison on SDD (KolektorSDD).}}
    \label{fig:app-qual-sdd}
\end{figure}

\begin{figure}[H]
    \centering
    \includegraphics[width=0.76\textwidth]{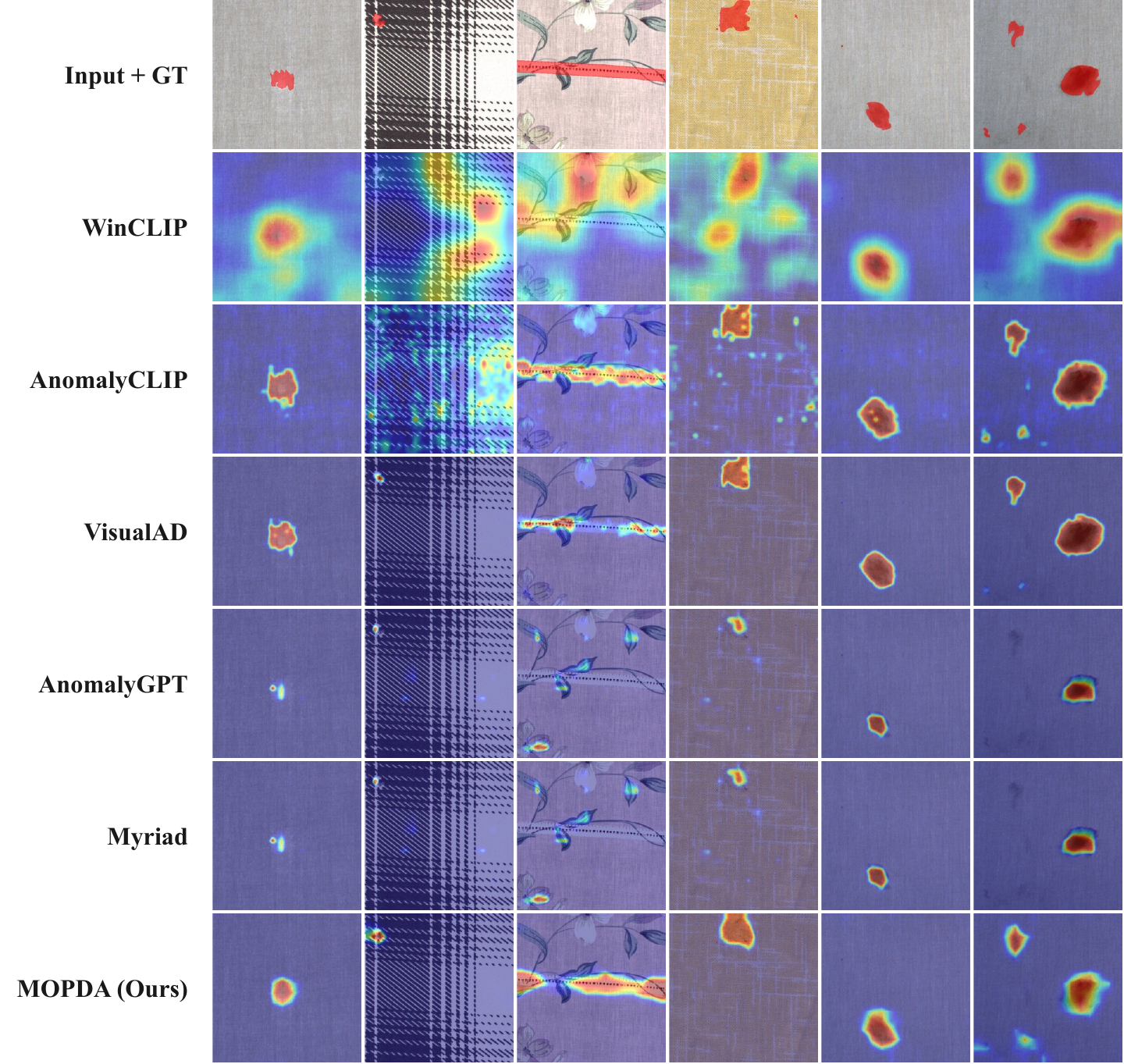}
    \caption{\textbf{Qualitative comparison on WFDD.}}
    \label{fig:app-qual-wfdd}
\end{figure}

\begin{figure}[H]
    \centering
    \includegraphics[width=0.76\textwidth]{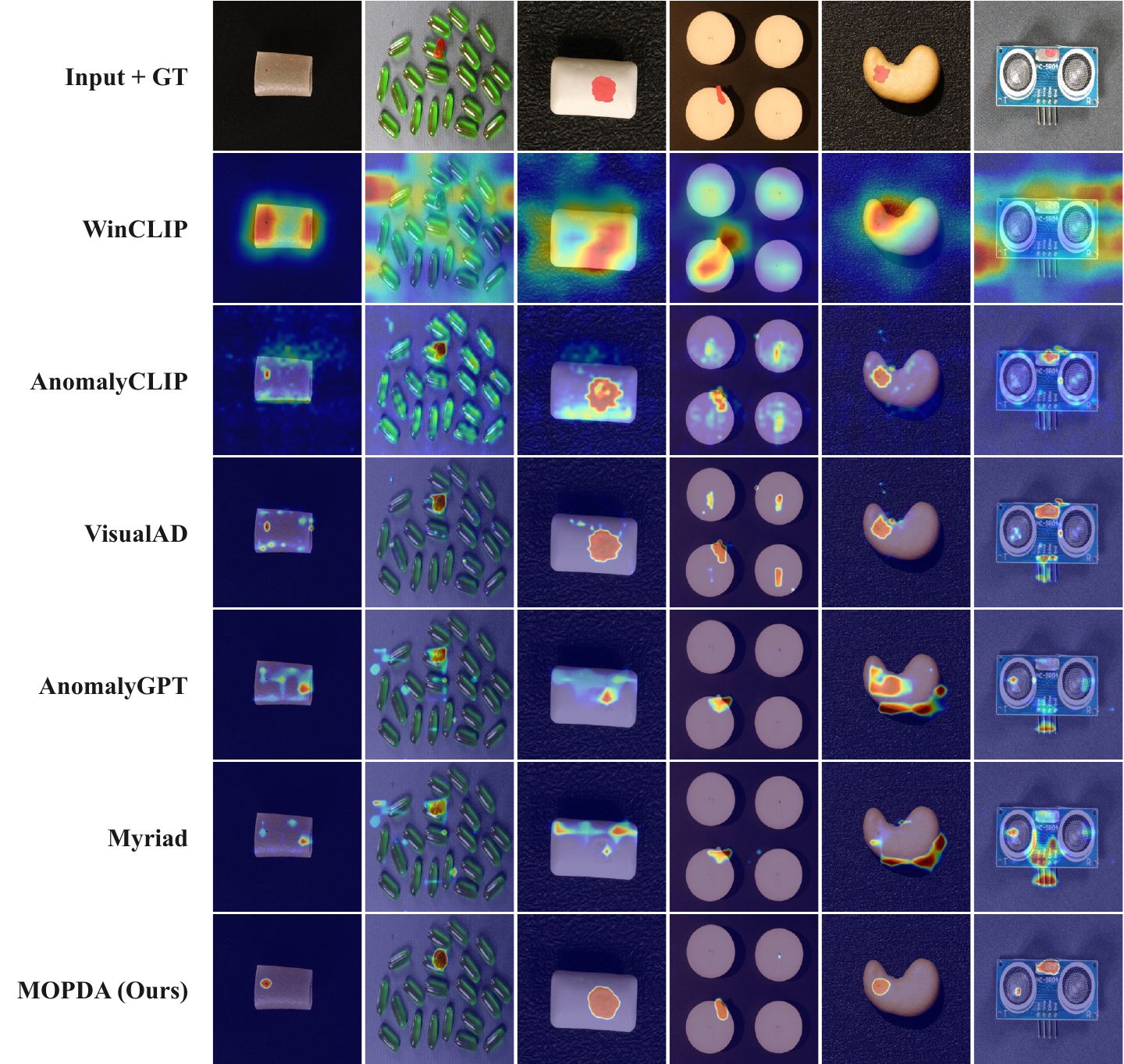}
    \caption{\textbf{Qualitative comparison on VisA.}}
    \label{fig:app-qual-visa}
\end{figure}

\begin{figure}[H]
    \centering
    \includegraphics[width=0.76\textwidth]{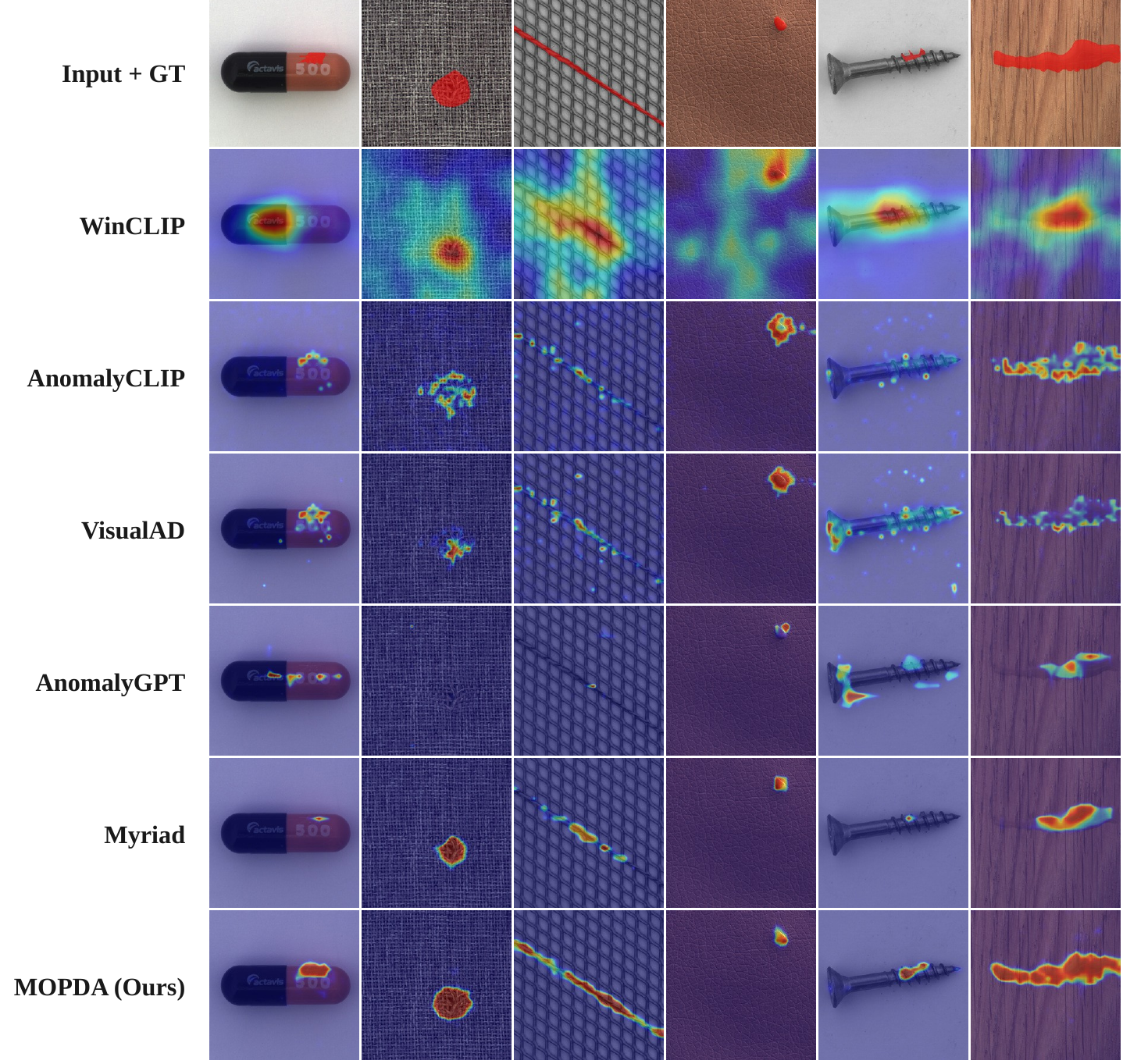}
    \caption{\textbf{Qualitative comparison on MVTec-AD.}}
    \label{fig:app-qual-mvtec}
\end{figure}

\begin{figure}[H]
    \centering
    \includegraphics[width=0.76\textwidth]{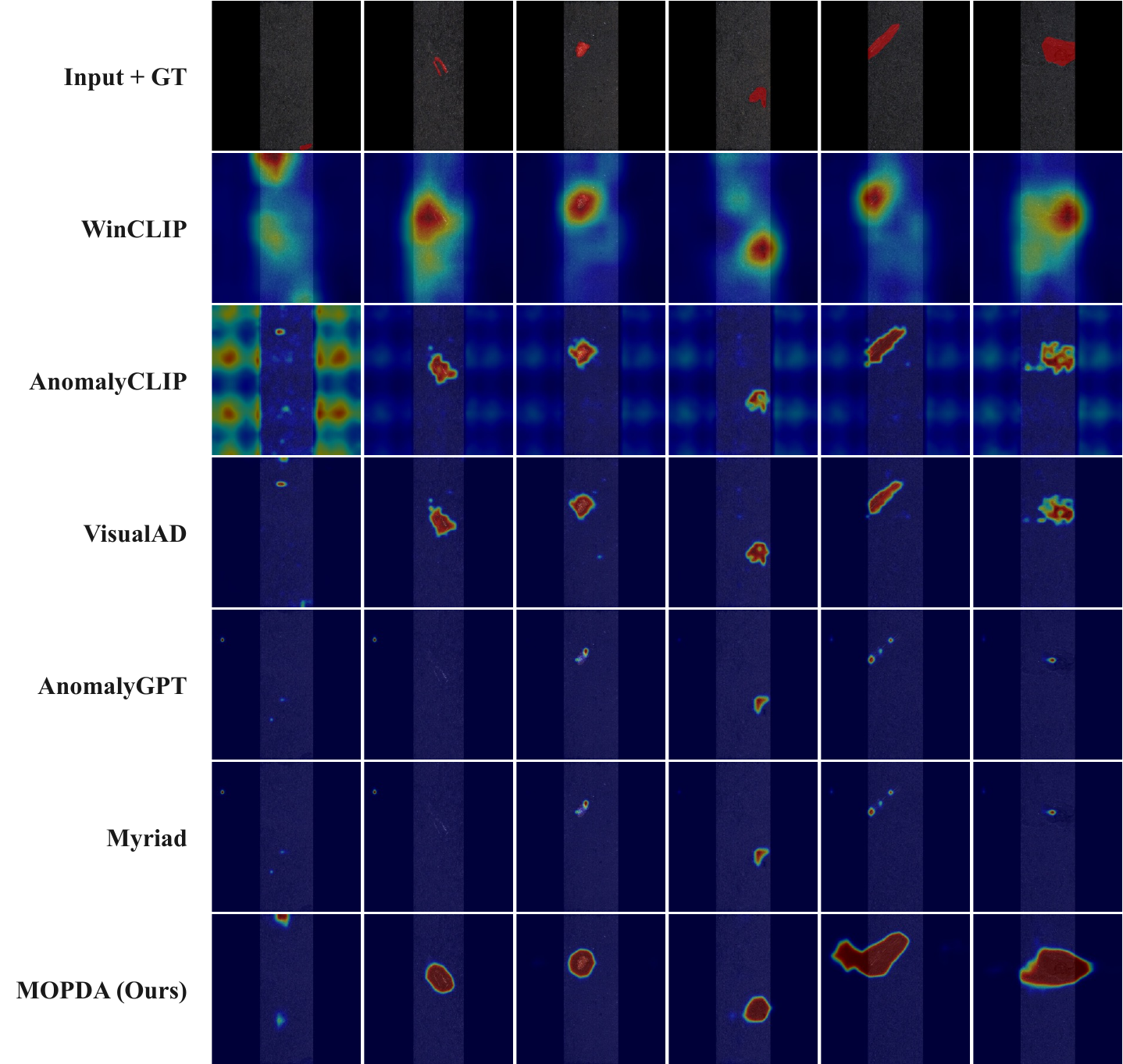}
    \caption{\textbf{Qualitative comparison on KSDD2.}}
    \label{fig:app-qual-ksdd2}
\end{figure}

\paragraph{Heatmap localization comparison.}
Figures~\ref{fig:app-qual-sdd}--\ref{fig:app-qual-ksdd2} use the same layout:
columns are target-domain samples, while rows show the input with its
ground-truth annotation followed by anomaly maps from WinCLIP, AnomalyCLIP,
VisualAD, AnomalyGPT, Myriad, and \method. The examples span object, texture,
and surface categories with compact, elongated, and irregular defects.

\paragraph{Thin structural defects on SDD.}
In Figure~\ref{fig:app-qual-sdd}, the baselines retain isolated fragments or
activate off-defect regions around narrow discontinuities. \method follows the
annotated defect more continuously and preserves its elongated structure. Its
heatmap extends beyond the binary annotation in some regions.

\paragraph{Fabric defects on WFDD.}
Figure~\ref{fig:app-qual-wfdd} contains compact marks and extended defects on
strongly textured cloth. The baselines often split their responses or follow
competing texture patterns, whereas \method concentrates on compact defects and
tracks the long horizontal defect, indicating transfer beyond source-domain
appearance.

\paragraph{Multi-instance objects on VisA.}
Figure~\ref{fig:app-qual-visa} tests scenes containing several similar objects.
\method places its strongest responses on the damaged capsules, candle, and
macaroni and follows their corrupted parts, whereas the baseline maps more
often spread across intact instances or unrelated regions.

\paragraph{Object and texture categories on MVTec-AD.}
Figure~\ref{fig:app-qual-mvtec} combines compact object defects with elongated
texture anomalies. \method remains localized on the capsule, carpet, leather,
and screw while retaining the long spatial support of the grid and wood
defects.

\paragraph{Surface defects on KSDD2.}
Figure~\ref{fig:app-qual-ksdd2} includes surface defects of different
sizes and shapes. \method covers a larger portion of several annotated
regions, whereas the baselines often emphasize isolated fragments.
Small and boundary-adjacent defects remain challenging, and some responses
extend beyond the annotated regions.

\paragraph{Cross-dataset observation.}
Across Figures~\ref{fig:app-qual-sdd}--\ref{fig:app-qual-ksdd2}, the examples
illustrate how localization behavior varies across target domains.
\method captures the spatial extent of many annotated defects, while
missed defects and overly broad responses remain visible in some cases.

\subsection{Judgment and Localization Case Studies}
\label{app:judgment-localization-cases}

\begin{figure}[H]
    \centering
    \includegraphics[width=0.90\linewidth]{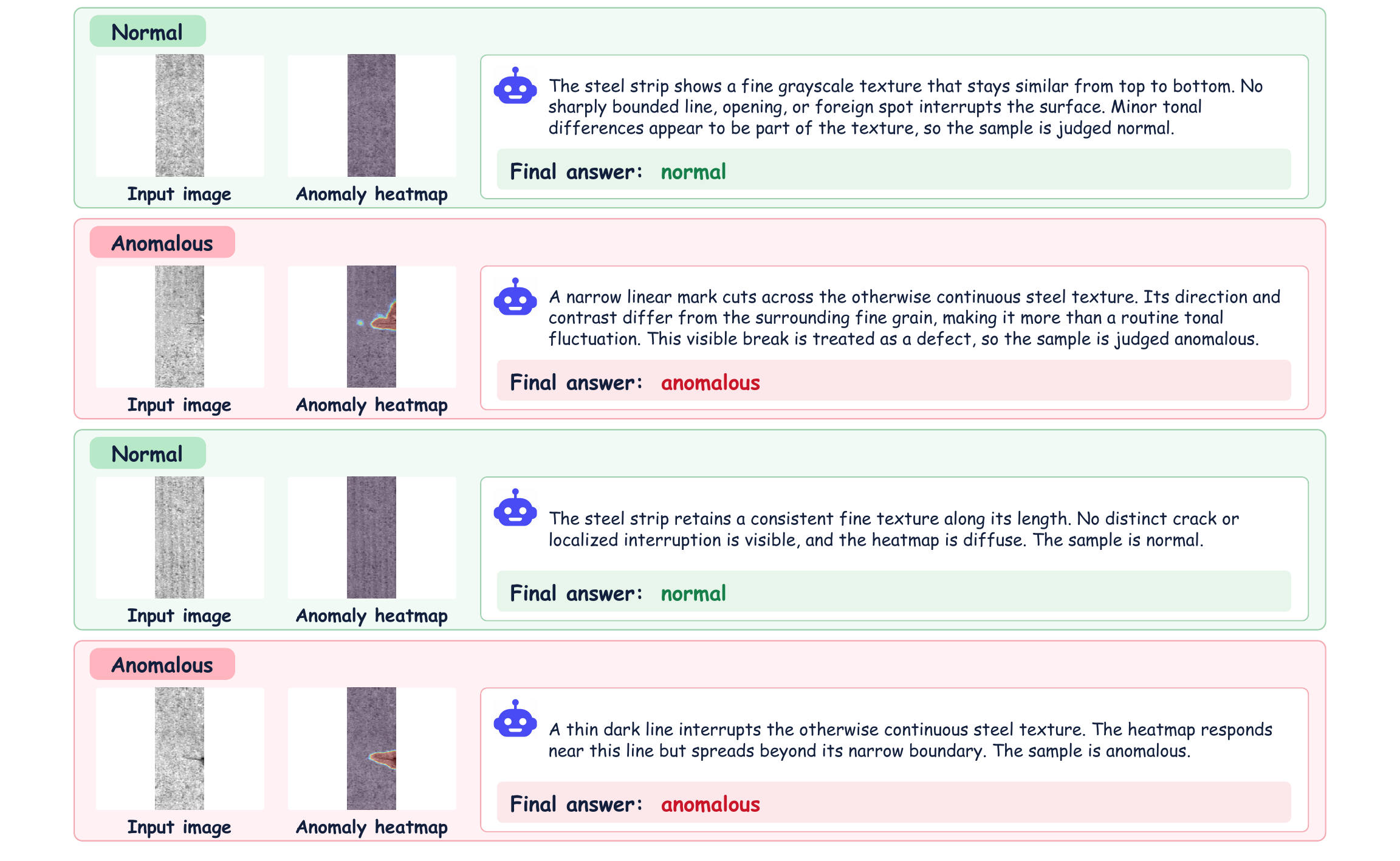}
    \caption{\textbf{Judgment and localization cases on SDD.}}
    \label{fig:app-judgment-localization-sdd}
\end{figure}

\begin{figure}[H]
    \centering
    \includegraphics[width=0.90\linewidth]{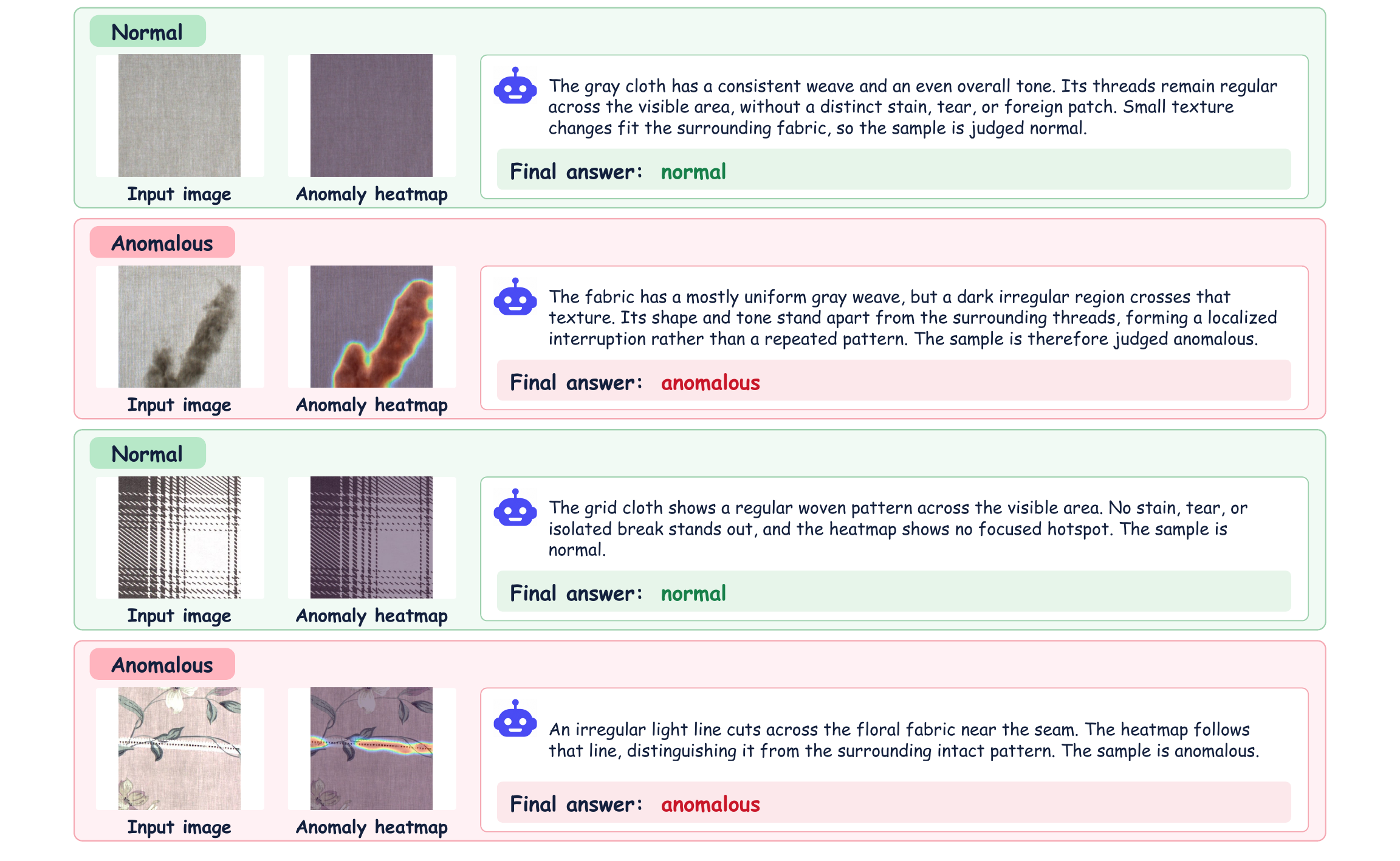}
    \caption{\textbf{Judgment and localization cases on WFDD.}}
    \label{fig:app-judgment-localization-wfdd}
\end{figure}

\begin{figure}[H]
    \centering
    \includegraphics[width=0.90\linewidth]{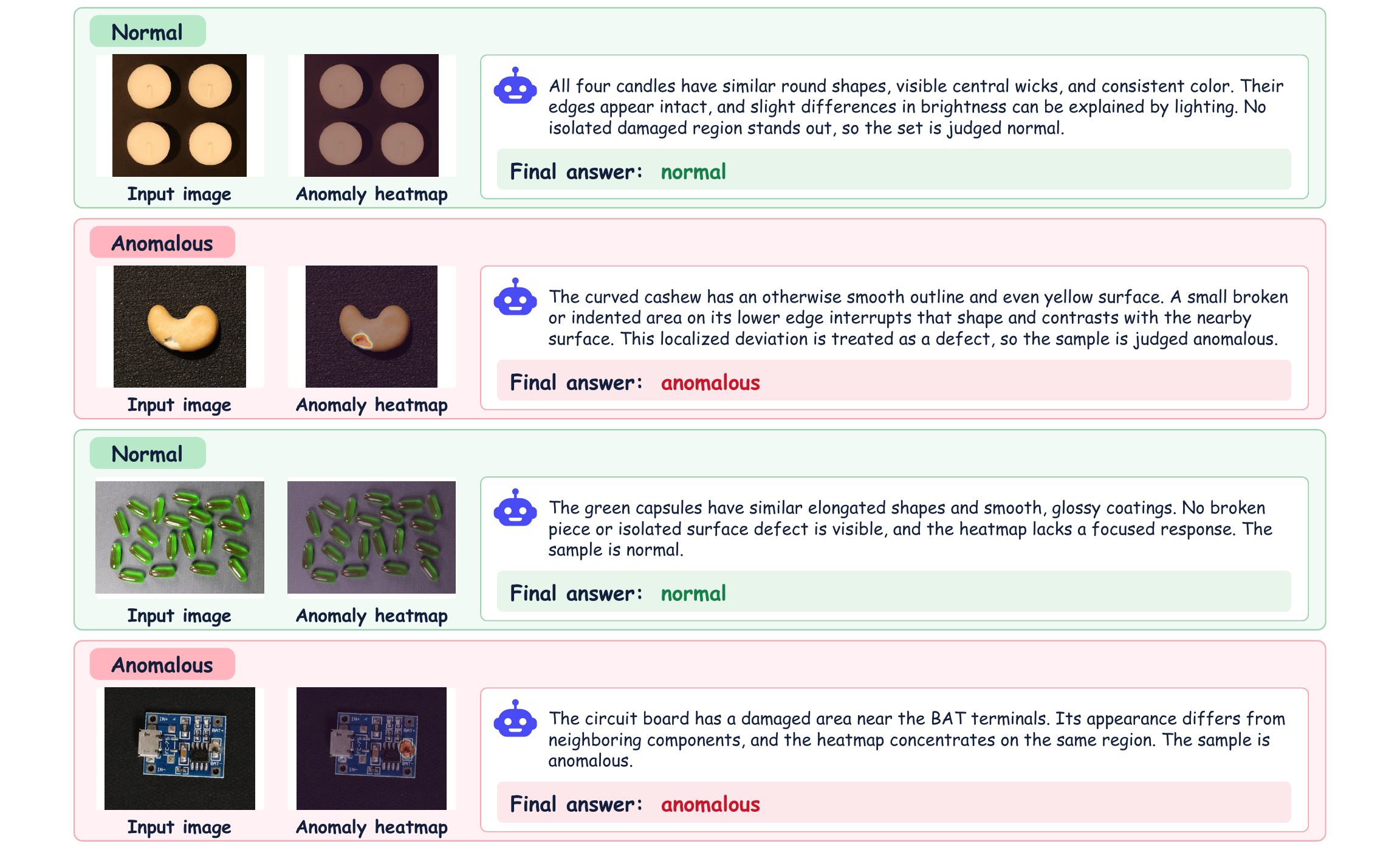}
    \caption{\textbf{Judgment and localization cases on VisA.}}
    \label{fig:app-judgment-localization-visa}
\end{figure}

\begin{figure}[H]
    \centering
    \includegraphics[width=0.90\linewidth]{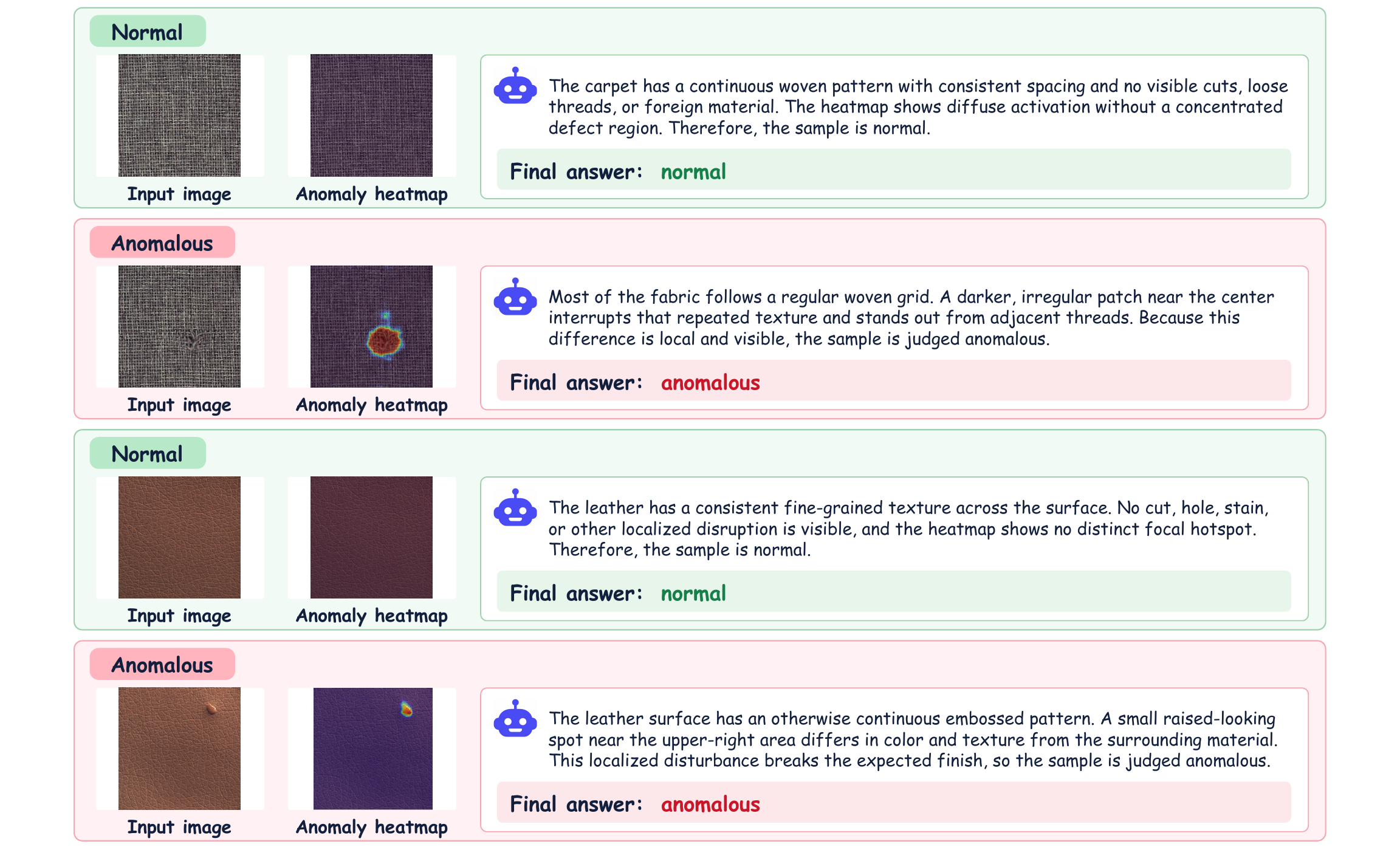}
    \caption{\textbf{Judgment and localization cases on MVTec-AD.}}
    \label{fig:app-judgment-localization-mvtec}
\end{figure}

\begin{figure}[H]
    \centering
    \includegraphics[width=0.90\linewidth]{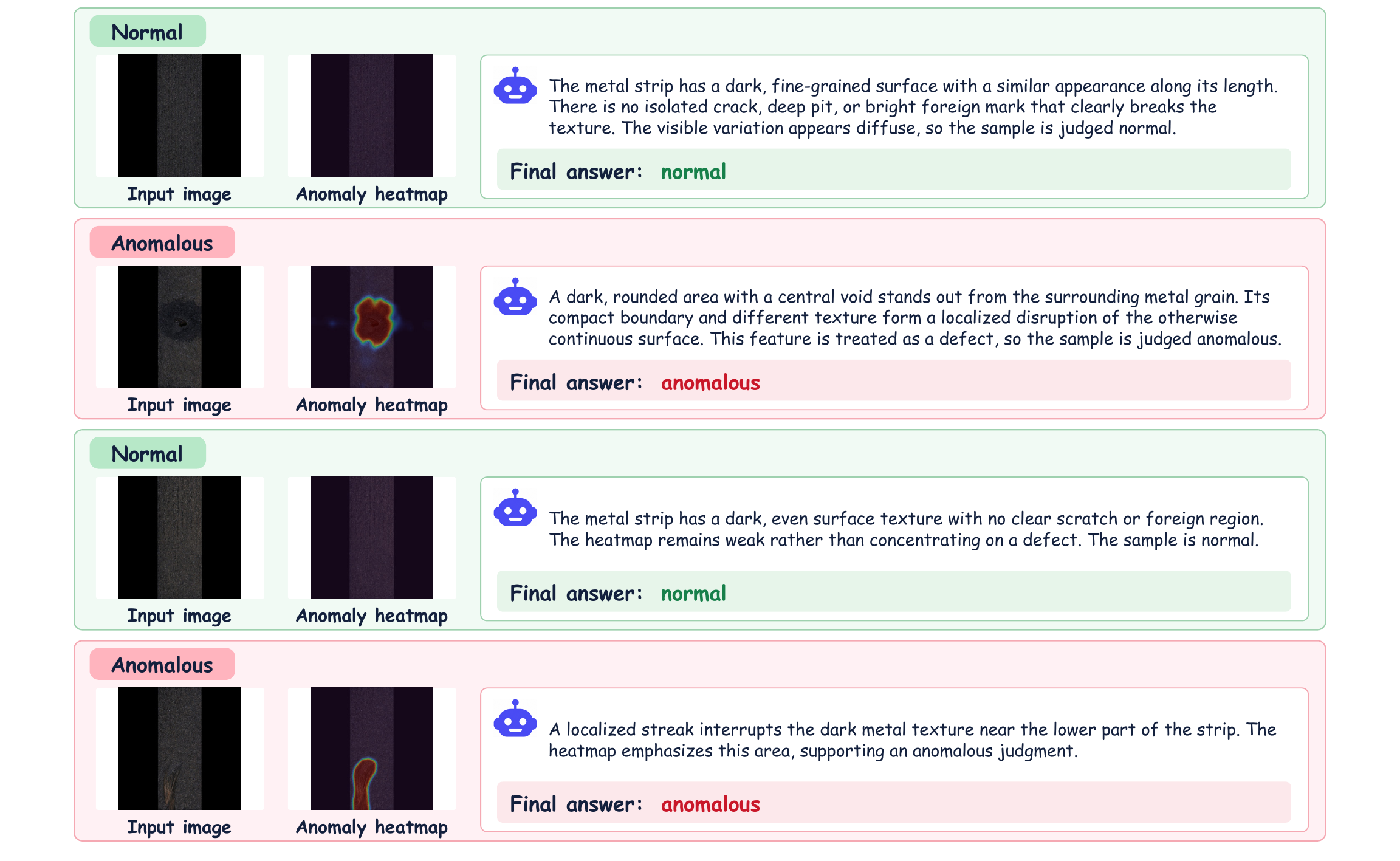}
    \caption{\textbf{Judgment and localization cases on KSDD2.}}
    \label{fig:app-judgment-localization-ksdd2}
\end{figure}

Figures~\ref{fig:app-judgment-localization-sdd}--
\ref{fig:app-judgment-localization-ksdd2} show four selected cases per
benchmark, with two normal and two anomalous samples in each. All selected
image-level judgments are correct. On normal samples, the anomaly maps remain
low-response overall and show no prominent spurious hotspots. On anomalous
samples, the strongest responses concentrate around visible defect regions.
These cases illustrate both low false-positive activation on normal surfaces
and localized responses to defects across target domains.

\subsection{Anchor-Space Visualization}
\label{app:anchor-space}

\begin{figure}[H]
\centering
\includegraphics[width=\textwidth]{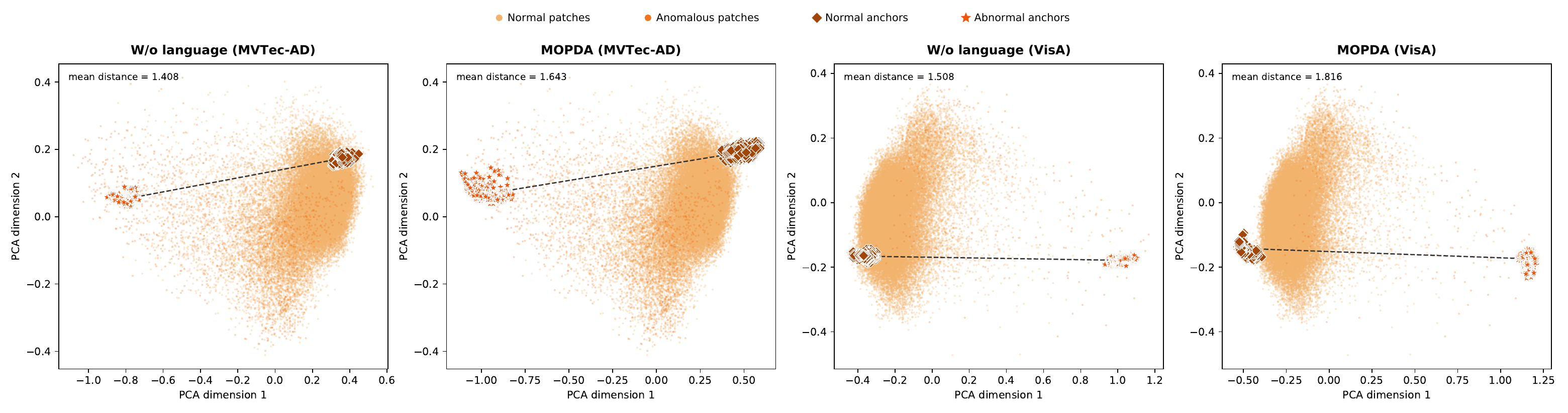}
\caption{\textbf{Anchor separation under final judgment.}
PCA visualization compares anchoring without language output and the final
\method condition on MVTec-AD and VisA. The final judgment increases the
normal-abnormal anchor distance and provides a clearer semantic direction for
visual-feature-grounded anomaly scoring.}
\label{fig:anchor-separation}
\end{figure}

Figure~\ref{fig:anchor-separation} visualizes the same effect in anchor space:
compared with anchoring without language output, the final-judgment condition
increases the separation between normal and abnormal anchors on both MVTec-AD
and VisA.

\section{Failure Cases and Limitations}
\label{app:limitations}

\begin{figure}[H]
\centering
\includegraphics[width=0.80\linewidth]{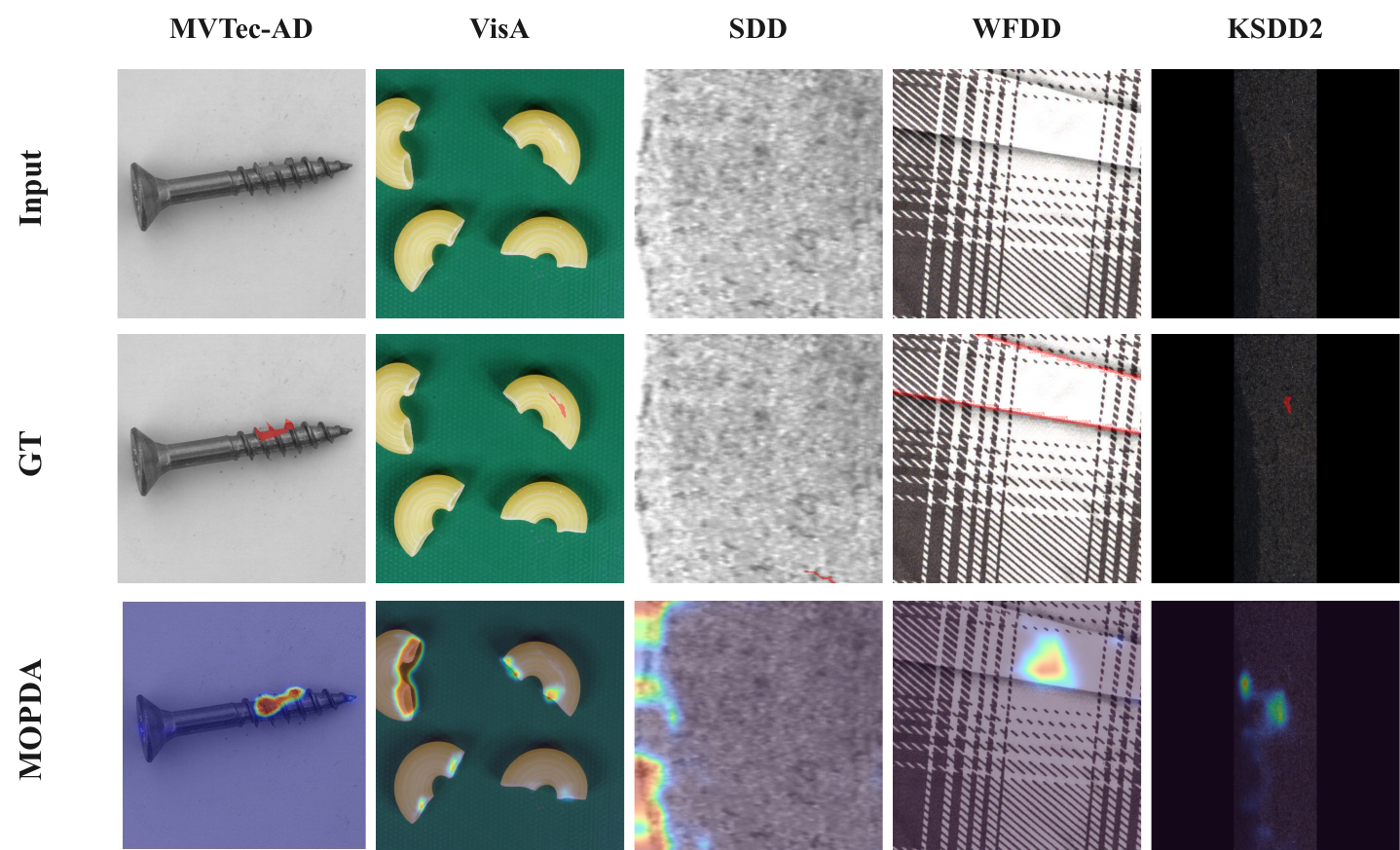}
\caption{\textbf{Localization failure cases.} Input, GT, and MOPDA response
for five benchmarks.}
\label{fig:app-localization-failures}
\end{figure}
\paragraph{Localization failures.}
Figure~\ref{fig:app-localization-failures} shows five selected cases in
which the predicted heatmap misses part of the annotated defect or activates
outside it.

\paragraph{Source-domain supervision.}
\method uses source-domain image labels, pixel masks, and reference answers
during training. Target-domain evaluation uses images without target annotations
or reference answers.

\paragraph{Binary response schema.}
The two-option answer format produces a stable normal/anomalous judgment for
visual anchoring. Richer response schemas could condition localization on defect
categories or calibrated uncertainty.

\end{document}